\documentclass[journal]{IEEEtai}

\PassOptionsToPackage{
  colorlinks=true,
  linkcolor=blue,
  citecolor=blue,
  urlcolor=blue,
  hypertexnames=false
}{hyperref}

\usepackage[numbers,square,sort&compress]{natbib}

\usepackage{bibunits}
\usepackage{chapterbib}
\usepackage{docmute}

\defaultbibliography{bibliography}
\defaultbibliographystyle{IEEEtran}

\usepackage{amsmath,amsfonts,amssymb}
\usepackage{bm}

\usepackage{algorithmic}
\usepackage{algorithm}

\usepackage{array}
\usepackage{url}
\usepackage{verbatim}
\usepackage{graphicx}
\usepackage[dvipsnames]{xcolor}
\usepackage{multirow}
\usepackage{colortbl}
\usepackage{orcidlink}
\usepackage{balance}
\usepackage{soul}
\usepackage{afterpage}
\usepackage{pdflscape}
\usepackage{longtable}
\usepackage{rotating}
\usepackage{pgf}
\usepackage{import}
\let\labelindent\relax
\usepackage{enumitem}

\usepackage{hyperref}

\hypersetup{
  colorlinks=true,
  linkcolor=blue,
  citecolor=blue,
  urlcolor=blue,
  hypertexnames=false
}

\makeatletter
\newcommand{\prefixcitationlinks}[1]{%
  \let\unprefixed@hyper@natlinkstart\hyper@natlinkstart
  \let\unprefixed@hyper@natlinkbreak\hyper@natlinkbreak
  \let\unprefixed@hyper@natanchorstart\hyper@natanchorstart
  \renewcommand{\hyper@natlinkstart}[1]{%
    \unprefixed@hyper@natlinkstart{#1##1}}%
  \renewcommand{\hyper@natlinkbreak}[2]{%
    \unprefixed@hyper@natlinkbreak{##1}{#1##2}}%
  \renewcommand{\hyper@natanchorstart}[1]{%
    \unprefixed@hyper@natanchorstart{#1##1}}%
}
\makeatother

\begin{document}


\begin{bibunit}[IEEEtran]

\title{Grad-CAM for Vision Transformers: A Systematic Taxonomy and Audit of Methodological Ambiguity in Explainable AI}

\author{Casey Wall\,\orcidlink{0009-0000-9973-6007}, 
Longwei Wang\,\orcidlink{0009-0002-0638-5637}, \textit{Member, IEEE,} 
Rodrigue Rizk\,\orcidlink{0000-0002-4392-4188}, \textit{Member, IEEE,} 
KC Santosh\,\orcidlink{0000-0003-4176-0236}
\textit{Senior, IEEE} 

    \thanks{This work was supported by the National Science Foundation under Grant No. \href{https://www.nsf.gov/awardsearch/showAward?AWD_ID=2346643}{\#2346643}, the U.S. Department of Defense under Award No. \href{https://dtic.dimensions.ai/details/grant/grant.14525543}{\#FA9550-23-1-0495}, and the U.S. Department of Education under Grant No. P116Z240151.
Any opinions, findings, conclusions or recommendations expressed in this material are those of the author(s) and do not necessarily reflect the views of the National Science Foundation, the U.S. Department of Defense, or the U.S. Department of Education.}

    \thanks{C. Wall, L. Wang, R. Rizk, and KC Santosh are with \href{https://ai-research-lab.org}{USD Artificial Intelligence Research}, Department of Computer Science, University of South Dakota (e-mail: casey.wall@coyotes.usd.edu, \{longwei.wang, rodrigue.rizk, kc.santosh\}@usd.edu).}
}


\maketitle

\begin{abstract}

Gradient-weighted Class Activation Mapping (Grad-CAM) is widely used to visualize model decisions, but it was originally formulated for convolutional neural networks, where spatial feature maps and channel dimensions have clear architectural meanings. Vision Transformers (ViTs) do not provide the same structure, instead representing images through tokens, attention, residual streams, and multimodal interactions. This paper presents a systematic taxonomy and literature audit of how Grad-CAM and related methods are adapted, justified, and reported for ViT-based architectures. From an initial search of more than 550 papers, we identify 175 papers that apply Grad-CAM or Grad-CAM-adjacent methods to ViTs. We find that most papers do not provide a full mathematical or implementation-level account of how Grad-CAM is adapted to transformer representations. To characterize this gap, we introduce a descriptive taxonomy of ViT Grad-CAM adaptations that makes explicit the feature locations, gradient targets, spatial reconstruction steps, and aggregation choices that are often left implicit. This taxonomy is not intended to prescribe a single correct adaptation, but to clarify the range of methodological choices being made. The study shows that Grad-CAM on ViTs is often treated as a trivial extension of CNN-based Grad-CAM, despite requiring nontrivial choices that affect rigor, reproducibility, and interpretation.
\end{abstract}

\begin{IEEEImpStatement}
This work supports more reliable and transparent use of explainable AI methods for vision transformer systems. Grad-CAM is often used to support claims about model behavior, trustworthiness, and visual grounding, including in high-stakes domains. However, when authors do not specify the feature representation, gradient target, token handling, aggregation strategy, or spatial reconstruction procedure, the resulting heatmap may be difficult to reproduce or evaluate. By auditing current reporting practices and providing a taxonomy of adaptation choices, this paper aims to improve methodological clarity rather than prescribe a single correct explanation method. Its broader impact is to help authors, reviewers, and readers distinguish reproducible attribution procedures from visually persuasive but underspecified explanations.

\end{IEEEImpStatement}


\section{Introduction}\label{sec:introduction}

\IEEEPARstart{G}{radient-weighted} Class Activation Mapping (Grad-CAM)~\cite{grad_cam} has become a widely used technique for visualizing the decision-making process of convolutional neural networks (CNNs) \cite{wang2019representation,wang2014congestion,wang2024enhanced,wang2018partial,wang2016optimization,wang2021explaining,wang2011exploration,shi2019deep,wang2017performance,wang2017low,wang2021improving,xiao2022looking,wang2016large,wang2011collaborative,wang2024dense,wang2019information,wang2016large,nayyem2024bridging,wang2024enhancing,wang2025explainability,ranabhat2025multi,uddin2025expert,rasmussen2025ecologically,chataut2024shape,wang2026expert,wang2025explainability,khadka2025coswin,wang2025bridging,rasmussen2026channel,wall2026winsor,ranabhat2025promoting,uddin2026learning,uddin2026explainable,wagle2026mechanistic,ranabhat2026frequency, wang2025explainability1, wang2016large1} . However, the recent shift toward Vision Transformers (ViTs)~\cite{ViT} and other token-based architectures introduces a fundamental mismatch: these models do not produce the convolutional feature maps or channel-wise activation patterns that Grad-CAM relies on. Despite this lack of theoretical grounding, Grad-CAM is routinely applied to ViTs across a growing body of literature, often without justification, adaptation, or acknowledgment of the architectural differences. This disconnect between the method's assumptions and its widespread use raises important questions about the validity, interpretability, and scientific rigor of such explanations.In this paper, we present a systematic taxonomy and literature audit of Grad-CAM use on ViTs, examining how often the method is applied, how authors justify its adaptation, and where methodological ambiguity remains.

This disconnect is not just a technical oversight; it reflects a broader issue in the explainable AI wherein methods are often applied in a ``plug-and-play'' manner without sufficient consideration of their underlying assumptions or the specific characteristics of the models they are being applied to. The result of this is a growing body of literature claiming to explain vision-attention-based model decisions in a manner that is not supported by the theoretical foundations of the method being used. Despite this widespread practice, there has been no systematic effort to define what it means to ``use Grad-CAM on ViTs,'' to evaluate the justifications provided, or to assess the implications of this methodological mismatch.

To address this gap, we conduct a systematic taxonomy and audit of papers that apply Grad-CAM to ViTs and related token-based architectures used in computer vision. Our study examines how frequently this practice occurs, how it is justified (if at all), how it is implemented, and what claims are made based on the resulting visualizations. We identify common patterns of misuse, misinterpretation, and over-claiming, as well as the risks associated with these practices for the field of explainable AI. Based on our findings, we provide recommendations for authors, reviewers, and the research community to promote more rigorous, architecture-aware approaches to explainability in vision models.

\section{Background and Taxonomy}\label{sec:background}

\subsection{Grad-CAM and Its Theoretical Assumptions}\label{sec:gradcam_background}

At basic level, Grad-CAM~\cite{grad_cam} requires spatial feature maps with height, width, and channel dimensions, where each channel encodes a meaningful visual pattern and where gradients of the target output with respect to these channels can be pooled into importance weights. This formulation assumes that each filter (or channel) contain independently meaningful visual patterns, and that, when linearly combined with gradient-based importance weights, produce a spatial heatmap that localizes output-relevant regions in the input image. The method relies on the hierarchical, spatially structured nature of convolutional feature maps, where each channel corresponds to a specific visual deep pattern or concept that can be weighted and combined to produce a class-discriminative localization map. However, ViTs and other token-based architectures do not produce such spatial feature maps or channel-wise activations. Instead, they generate token embeddings that are processed through self-attention mechanisms, which do not have the same spatial structure or interpretability as convolutional feature maps. As a result, applying Grad-CAM to ViTs involves a fundamental mismatch between the method's assumptions and the model's architecture, leading to visualizations that may not be meaningful or interpretable in the same way as they are for CNNs.

\subsubsection{Grad-CAM Recap}\label{sec:gradcam_recap}

Grad-CAM~\cite{grad_cam} is a widely used technique for visualizing class-specific discriminative regions in CNNs. Given a target class $c$, Grad-CAM computes the gradient of the class score $y^c$ with respect to the activation maps of a selected convolutional layer. For a feature map $A^k \in \mathbb{R}^{H \times W}$, the importance weight for the $k$-th channel is defined as
\[
\alpha_k^c \;=\; \frac{1}{H W} 
\sum_{u=1}^{H}\sum_{v=1}^{W}
\frac{\partial y^c}{\partial A^k_{uv}},
\]
where $A^k_{uv}$ denotes the activation at spatial location $(u,v)$ and $H$ and $W$ are the spatial dimensions of the feature map. These channel-wise importance weights are then used to form a class-specific localization map by a weighted combination of the feature maps, followed by a ReLU activation:

\[
L^c_{\text{Grad-CAM}} 
= \mathrm{ReLU}\!\left( \sum_k \alpha_k^c A^k \right),
\]
yielding a heatmap $L^c_{\text{Grad-CAM}} \in \mathbb{R}^{H \times W}$ that highlights the spatial regions most influential for predicting class $c$. The resulting heatmap is typically upsampled to the input resolution and normalized for visualization.

\subsection{Vision Transformers and Token-Based Architectures}\label{sec:vit_background}

Before analyzing how Grad-CAM has been applied to ViTs in the literature, it is necessary to make explicit what it means to choose a ``feature map'' in a transformer architecture. In CNNs, Grad-CAM is naturally defined over convolutional activation maps, where spatial locations and channel dimensions have clear architectural meanings. In ViTs, however, there is no single equivalent object. Depending on the implementation, a Grad-CAM adaptation may use patch-token embeddings, attention matrices, attention outputs, multi-layer perceptron (MLP) activations, residual-stream representations, or cross-attention maps in multimodal models. Each choice changes the mathematical meaning of the resulting heatmap.

To address this ambiguity, we develop a taxonomy of possible Grad-CAM feature extraction locations and aggregation strategies for ViT-based architectures. This taxonomy is not intended to introduce a new explanation method. Rather, it provides a framework for describing and comparing the many ways existing papers implicitly adapt Grad-CAM to transformer models. To the best of our knowledge, prior work has not provided a unified taxonomy that systematically distinguishes between token-feature methods, attention-map methods, head-wise attribution methods, residual-stream feature choices, and cross-attention variants in Vision Language Models (VLMs). Establishing this taxonomy allows us to evaluate whether later papers specify enough methodological detail to justify their use of Grad-CAM on ViTs.

\subsubsection{ViTs and the rise of attention-based vision models}

The introduction of ViTs~\cite{ViT} marked a significant shift in the field of computer vision, moving away from convolutional architectures toward transformer-based models that rely on token-based representations. ViTs process images by dividing them into fixed-size patches, which are then linearly embedded and treated as tokens. Mathematically this can be represented as follows: given an input image $I \in \mathbb{R}^{H \times W \times C}$, it is divided into $N$ patches of size $P \times P$, these patches are then flattened and projected into a $D$-dimensional embedding space that is positionally encoded to produce a sequence of token embeddings $X \in \mathbb{R}^{N \times D}$. These token embeddings are then processed through multiple layers of multi-head self-attention and feedforward networks, which do not produce spatial feature maps or channel-wise activations in the same way as CNNs. The attention mechanism allows the model to capture long-range dependencies and global context. While the attention mechanism provides a powerful way to model relationships between different parts of the input, it does not inherently produce the kind of spatially localized feature maps, rather it relies on inter-patch relationships and positional encodings to capture spatial information.

\begin{figure*}[t]
    \centering
    \includegraphics[width=0.98\textwidth]{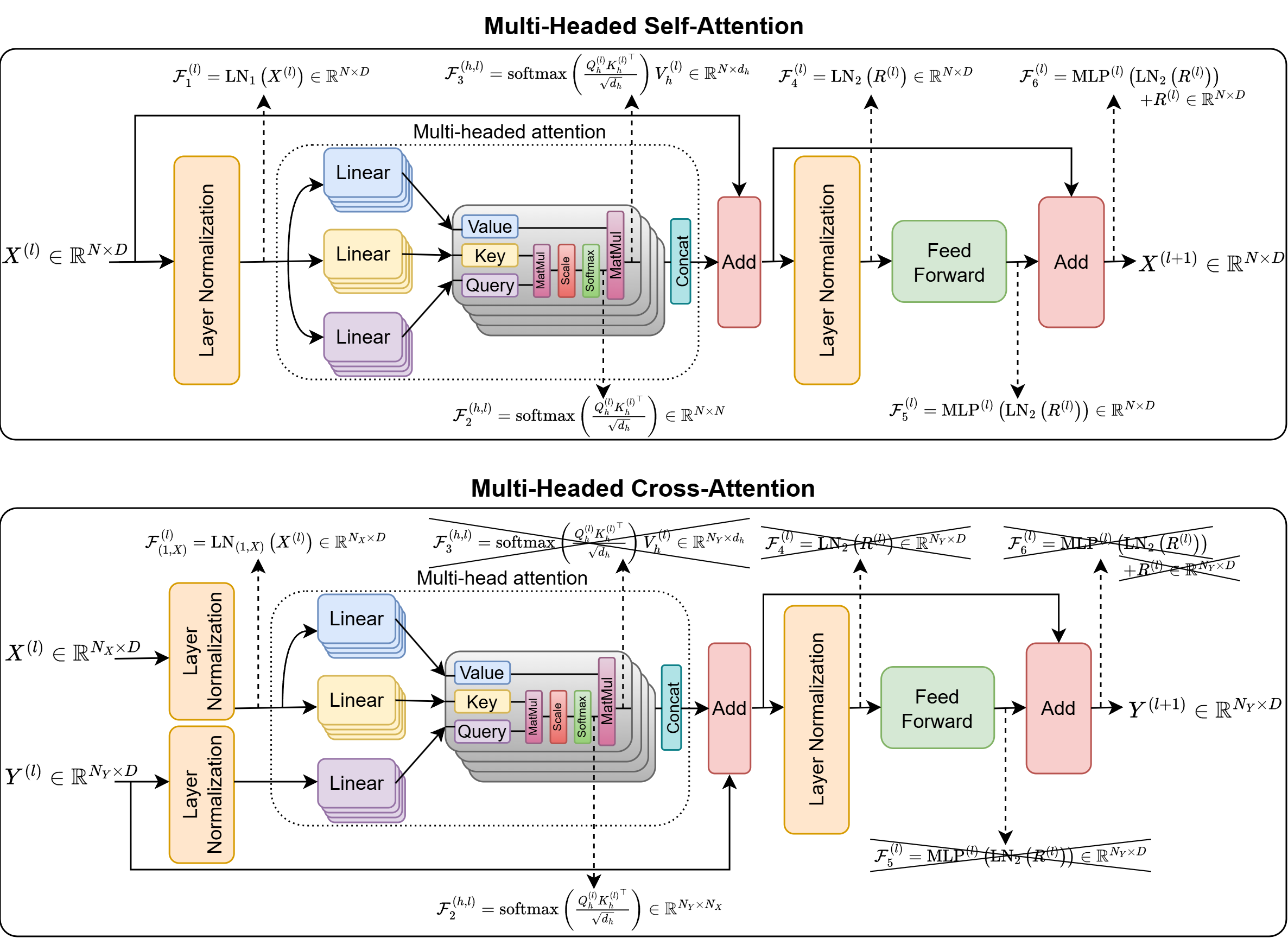}
    \caption{Feature extraction locations for adapting Grad-CAM to transformer-based vision models. The top panel shows a standard ViT self-attention encoder block, where token-feature representations, attention maps, attention outputs, MLP activations, and residual outputs may all be considered possible attribution sources. The bottom panel shows a cross-attention block commonly used in vision-language models, where the query sequence is assumed to come from a non-spatial modality such as text and the key/value sequence is derived from visual tokens. Under this assumption, the cross-attention matrix $\mathcal{F}_2$ provides the clearest image-token attribution source, while later cross-attention outputs $\mathcal{F}_3$--$\mathcal{F}_6$ are crossed out because they are text-side representations and do not directly preserve a spatial visual-token layout.}
    \label{fig:vit_grad_cam_features}
\end{figure*}

\subsubsection{Feature extraction locations in ViTs}\label{sec:feature_extraction_locations}

In ViTs there are several locations within the transformer encoder architecture from which feature representations analogous to the activation maps, \(A\), used in CNN-based Grad-CAM~\cite{grad_cam} methods may be extracted. These representations originate from different stages of the transformer block and capture distinct forms of information, including patch-token representations, attention relationships, and globally contextualized embeddings. Fig.~\ref{fig:vit_grad_cam_features} illustrates several common extraction locations used for Grad-CAM-style visualizations in ViT architectures. We mainly focus on the standard trans former encoder block structure as it is defined in~\cite{ViT}, but the same principles apply to other ViT-based architectures that rely on tokenized representations and attention mechanisms.

One of the most commonly used extraction locations is the output of the first layer normalization (LN) operation applied prior to the multi-head attention module, which corresponds to the normalized token embeddings. This representation may be written as
\begin{equation}
\mathcal{F}_1^{(l)}
=
\mathrm{LN}_1\left(X^{(l)}\right)
\in \mathbb{R}^{N \times D},\label{eq:feature_extraction_pre_attention_ln}
\end{equation}
where \(X^{(l)}\) denotes the input token embeddings to the \(l\)-th transformer block, \(\mathrm{LN}_1\) denotes the first layer normalization operation, \(N\) is the number of input tokens, and \(D\) is the embedding dimension. Since this representation exists prior to attention-based information aggregation, it does not take into consideration inter-patch relationships or global context. Instead, the layer normalization operates strictly on intra-patch statistics, calculating per-patch mean and variance while applying learned channel-wise scale and bias parameters. Nonetheless, this extraction location is commonly used in ViT Grad-CAM implementations, including the \texttt{pytorch-grad-cam} library~\cite{jacobgil2021pytorchgradcam}, as well as in numerous studies applying Grad-CAM to transformer-based vision models.

A second candidate feature representation for ViT Grad-CAM is the softmax-normalized attention matrix produced within the multi-headed attention module. This representation may be expressed as
\begin{equation}
\mathcal{F}_{2}^{(h,l)}
=
\mathrm{softmax}
\left(
\frac{Q_h^{(l)} {K_h^{(l)}}^\top}
{\sqrt{d_h}}
\right)
\in \mathbb{R}^{N \times N},\label{eq:feature_extraction_attention_map}
\end{equation}
where $Q_h^{(l)} \in \mathbb{R}^{N \times d_h}$ and $K_h^{(l)} \in \mathbb{R}^{N \times d_h}$ are the query and key matrices associated with attention head \(h\) in transformer block \(l\), and \(d_h\) denotes the dimensionality of the attention head. Unlike conventional CNN feature maps, these attention matrices primarily encode token-to-token interaction strengths rather than localized semantic activations. Nevertheless, they are widely used in transformer explainability methods because they explicitly characterize how tokens attend to one another. This is particularly useful in multimodal vision-language transformers, where attention maps can visualize interactions between image and text tokens within cross-attention modules, although they are also frequently employed in single-modal ViT architectures.

Feature representations may also be extracted from the contextualized token representations produced by individual attention heads prior to output projection. This representation is given by
\begin{equation}
\mathcal{F}_{3}^{(h,l)}
=
\mathrm{softmax}
\left(
\frac{Q_h^{(l)} {K_h^{(l)}}^\top}
{\sqrt{d_h}}
\right)
V_h^{(l)}
\in \mathbb{R}^{N \times d_h},\label{eq:feature_extraction_attention_output}
\end{equation}
where \(V_h^{(l)}\in \mathbb{R}^{N \times d_h}\) denotes the value matrix associated with attention head \(h\). This tensor corresponds to the output of an individual attention head before concatenation across heads and application of the output projection matrix.

Although $\mathcal{F}_{3}^{(h,l)}$ is not itself a spatial saliency map, its first dimension remains indexed by token position. After removing the \texttt{[CLS]} token, the remaining (N-1) rows correspond directly to the spatial patch tokens used by the ViT. Grad-CAM weights may therefore be applied channel-wise to the patch-token activations, after which summation over the channel dimension produces a vector in ($\mathbb{R}^{N-1}$). This vector can then be reshaped into the original patch layout used by the model.

Because $\mathcal{F}_{3}^{(h,l)}$ is computed after the attention-weighted aggregation of the value matrix, each token representation already incorporates contextual information from other tokens. Consequently, the resulting visualizations may exhibit weaker spatial locality compared to representations that preserve stronger token-wise independence.

For the next representation let
\begin{equation}
R^{(l)}
=
X^{(l)}
+   
\mathrm{Concat}
\left(
\mathrm{head}_1^{(l)},
\dots,
\mathrm{head}_H^{(l)}  
\right)
W_O \in \mathbb{R}^{N \times D},\label{eq:residual_addition}
\end{equation}
denote the intermediate token representation after the first residual addition and output projection, but prior to the second learnable layer normalization and feedforward network. We would not typically use this representation for Grad-CAM visualizations as it is more analogous to a deep residual representation in a CNN rather than an activation map. Though it could still be still be theoretically be reshaped into valid feature maps.

Using the intermediate residual representation $R^{(l)}$, three additional candidate Grad-CAM feature tensors can be defined from the feedforward portion of the transformer block. The first is the token representation after the second layer normalization operation:

\begin{equation}
\mathcal{F}_{4}^{(l)}
=
\mathrm{LN}_2
\left(
R^{(l)}
\right)
\in \mathbb{R}^{N \times D},
\label{eq:feature_extraction_ffn_pre_residual_ln}
\end{equation}
where $\mathcal{F}_{4}^{(l)}$ denotes the normalized token representation passed into the feedforward network. The second is the output of the feedforward network itself:
\begin{equation}
\mathcal{F}_{5}^{(l)}
=
\mathrm{MLP}^{(l)}
\left(
\mathrm{LN}_2
\left(
R^{(l)}
\right)
\right)
\in \mathbb{R}^{N \times D},
\label{eq:feature_extraction_ffn_pre_residual}
\end{equation}
where $\mathrm{MLP}^{(l)}$ denotes the feedforward network within the $l$-th transformer block. Here, $\mathcal{F}_{5}^{(l)}$ represents the transformed token features before the second residual addition. Finally, the post-feedforward residual output can also be treated as a candidate feature tensor:
\begin{equation}
\mathcal{F}_{6}^{(l)}
=
R^{(l)}
+
\mathrm{MLP}^{(l)}
\left(
\mathrm{LN}_2
\left(
R^{(l)}
\right)
\right)
\in \mathbb{R}^{N \times D}.
\label{eq:feature_extraction_ffn}
\end{equation}
Thus, $\mathcal{F}_{4}^{(l)}$, $\mathcal{F}_{5}^{(l)}$, and $\mathcal{F}_{6}^{(l)}$ all denote token-indexed feature tensors from the feedforward half of the transformer block. They have the same basic shape as the patch-token representations used in other token-feature Grad-CAM variants, but they occur after additional normalization, nonlinear transformation, and residual mixing. For this reason, they are theoretically usable for Grad-CAM-style visualization after removing non-spatial tokens and reshaping the patch-token axis, but they may provide weaker spatial fidelity than earlier token representations.

Generally, the selection of feature extraction location is particularly important in ViTs because repeated self-attention operations progressively mix information across all tokens throughout the network depth. Consequently, deeper transformer representations often lose the strictly localized spatial structure naturally preserved within convolutional feature maps. Different extraction locations therefore emphasize different aspects of the learned representation, including token-level semantic content, token-to-token interactions, or globally contextualized feature representations, all of which can substantially affect the interpretability and spatial characteristics of Grad-CAM visualizations in transformer-based architectures.

\subsubsection{Reshaping problem and token ordering} \label{sec:reshaping_problem}

Standard Grad-CAM~\cite{grad_cam} formulations assume feature maps with an explicit spatial structure. In VITs, feature representations instead take the form of token embeddings $\mathcal{F}^{(l)} \in \mathbb{R}^{N \times D}$, where $N$ is the number of tokens and $D$ is the embedding dimension. Unlike CNN feature maps, these representations do not natively encode height and width structure.

However, since the sequence contains a structured ordering of image patches, the token dimension can be mapped back to a spatial grid after removing the \texttt{[CLS]} token. Specifically, if $N = \frac{H}{P} \times \frac{W}{P} + 1$, then the patch-token representation $\Tilde{\mathcal{F}}^{(l)} \in \mathbb{R}^{(N-1) \times D}$ can be reshaped into a grid of size $\frac{H}{P} \times \frac{W}{P}$, where the embedding dimension $D$ is treated as the channel dimension. Importantly, this reshaping is applied only for visualization and does not imply that the transformer internally represents spatial structure.

For attention-based feature maps such as $\mathcal{F}^{(h,l)}_2 \in \mathbb{R}^{N \times N}$ (attention matrices) and $\mathcal{F}^{(h,l)}_3 \in \mathbb{R}^{N \times d_h}$ (value-projected token representations), spatial structure is not explicitly encoded. Instead, spatial alignment must be recovered by operating on the token axis corresponding to image patches.

In practice, a common approach is to construct token-aligned representations by selecting interactions or activations associated with image patches and discarding the \texttt{[CLS]} token. This produces representations over patch tokens only, yielding structures in $\mathbb{R}^{N-1}$ that can subsequently be mapped back to a spatial grid using the known patch ordering induced by the ViT.

Heads are typically either aggregated or treated as separate feature channels prior to computing the final Grad-CAM weighting. In the simplest case, CLS-conditioned token representations are used as feature maps. Specifically, for multi-head attention-based feature maps, each head is treated as an independent channel, and only CLS-to-patch interactions (or CLS-conditioned activations, in the case of value features) are retained. This yields:

\begin{equation}
\hat{\mathcal{F}}^{(h,l)}_{2}
=
\mathcal{F}^{(h,l)}_{2}[\texttt{CLS}, 2:N]
\in \mathbb{R}^{N-1},
\end{equation}\label{reshaping_attention_map}
or
\begin{equation}
\hat{\mathcal{F}}^{(h,l)}_{3}
=
\mathcal{F}^{(h,l)}_{3}[\texttt{CLS}, 2:N]
\in \mathbb{R}^{d_h}.\label{reshaping_attention_value}
\end{equation}

These representations define token-aligned saliency vectors over image patches. Under the standard ViT patch ordering assumption, the index set $\{2,\dots,N\}$ corresponds to a rasterized arrangement of image patches, allowing these vectors to be reshaped into a spatial grid of size $\frac{H}{P} \times \frac{W}{P}$ for visualization. In this formulation, the embedding dimension (for $\mathcal{F}_3$) or head dimension (for $\mathcal{F}_2$-based constructions) serves as the channel axis in subsequent linear combinations with gradient-based importance weights.

The first approach uses only the gradients of the tokens corresponding to image patches, excluding the \texttt{[CLS]} token. In this case, the importance weights are computed as
\begin{equation}
\Tilde{\alpha}_k^{(l,c)} =
\frac{1}{N-1}
\sum_{i \in {2, \dots, N}}
\frac{\partial y^c}
{\partial \left[\mathcal{F}^{(l)}_m\right]_{i,k}},\label{token_gradients}
\end{equation}
where $m \in {1,4,5,6}$ denotes the selected feature-map extraction location. Here, $\Tilde{\alpha}_k^{(l,c)}$ represents the average importance weight for the $k$-th channel across all tokens corresponding to image patches, explicitly excluding the \texttt{[CLS]} token.

For multi-headed feature maps ($m \in {2,3}$), the importance weights are computed separately for each attention head. In this case, the gradients are again averaged over all tokens corresponding to image patches, excluding the \texttt{[CLS]} token. The importance weights are defined as
\begin{equation}
\Tilde{\alpha}_{h,k}^{(l,c)} =
\frac{1}{N-1}
\sum_{i \in {2, \dots, N}}
\frac{\partial y^c}
{\partial \left[\mathcal{F}^{(h,l)}_m\right]_{i,k}}.\label{token_gradients_heads}
\end{equation}
The second approach uses only the gradients from the \texttt{[CLS]} token to weight the selected feature map. In this case, the importance weights are computed as
\begin{equation}
\Hat{\alpha}_k^{(l,c)} =
\frac{\partial y^c}
{\partial \left[\mathcal{F}^{(l)}_m\right]_{1,k}},\label{cls_gradients}
\end{equation}
for $m \in {1,4,5,6}$, where index $1$ denotes the \texttt{[CLS]} token. For multi-headed feature maps ($m \in {2,3}$), the importance weights are computed separately for each attention head:
\begin{equation}
\Hat{\alpha}_{h,k}^{(l,c)} =
\frac{\partial y^c}
{\partial \left[\mathcal{F}^{(h,l)}_m\right]_{1,k}}.\label{cls_gradients_heads}
\end{equation}

\subsubsection{Linear combination and aggregation choices}

Finally, the linear combination can be performed in multiple ways. For non-multi-headed feature maps there are only two logical approaches to the linear combination step: either using the average token gradients as importance weights (eq.~\ref{token_gradients}) or using the \texttt{[CLS]} token gradients as importance weights (eq.~\ref{cls_gradients}). This can be formalated as follows:
\begin{equation}
L^{(l,c)}_{\text{Flat-GCAM}} = \text{ReLU} \left( \sum_k \omega_k^{(l,c)} \Tilde{\mathcal{F}}_m^{(l)} \right)\in \mathbb{R}^{N-1}\label{linear_combination_non_attention}
\end{equation}
where $\omega_k^{(l,c)}$ is either $\Tilde{\alpha}_k^{(l,c)}$ from eq.~\ref{token_gradients} or $\Hat{\alpha}_k^{(l,c)}$ from eq.~\ref{cls_gradients}, and $m\in\{1,4,5,6\}$.

For multi-headed feature representations, there are multiple valid ways to perform the linear combination step. Here, we use the term \emph{Flat-GCAM} to denote the one-dimensional token-level Grad-CAM attribution vector obtained before reshaping patch-token scores back into the two-dimensional image grid. In the multi-headed case, each attention head may either be combined before the ReLU operation or treated separately and aggregated after ReLU. For a selected multi-headed feature representation $\hat{\mathcal{F}}^{(h,l)}_m$, where $m\in{2,3}$, one possible flat attribution map is
\begin{equation}
L^{(l,c)}_{\text{Flat-GCAM}, i} =
\mathrm{ReLU} \left(\sum_h
\sum_k
\omega_{h,k}^{(l,c)}
\hat{\mathcal{F}}_{m,i,k}^{(h,l)}
\right),\label{linear_combination_attention_cls}
\end{equation}
or
\begin{equation}
L^{(l,c)}_{\text{Flat-GCAM}, i} =
\sum_h \mathrm{ReLU} \left(
\sum_k \omega_{h,k}^{(l,c)}
\hat{\mathcal{F}}_{m,i,k}^{(h,l)}
\right),\label{linear_combination_attention_cls_separate}
\end{equation}
where $\omega_{h,k}^{(l,c)}$ is either $\Tilde{\alpha}_{h,k}^{(l,c)}$ from eq.~\ref{token_gradients_heads} or $\Hat{\alpha}_{h,k}^{(l,c)}$ from eq.~\ref{cls_gradients_heads}, $m\in\{2,3\}$, and $\quad i \in \{2,\dots,N\}$.

Finally, to obtain a spatial heatmap for visualization, the Flat-GCAM representation is re-indexed according to the known patch ordering of the ViT. Since patch tokens theoretically correspond to a fixed raster scan of the input image, the vector $L^{(l,c)}_{\text{Flat-GCAM}} \in \mathbb{R}^{N-1}$ can be mapped to a 2D grid of size $\frac{H}{P} \times \frac{W}{P}$:
\begin{equation}
L^{(l,c)}_{\text{Grad-CAM}} =
\text{reshape}\left(
\{L^{(l,c)}_{\text{Flat-GCAM},i}\}_{i=1}^{N-1},
\left(\frac{H}{P}, \frac{W}{P}\right)
\right),\label{reshaping}
\end{equation}
where $L^{(l,c)}_{\text{Grad-CAM}} \in \mathbb{R}^{\frac{H}{P} \times \frac{W}{P}}$. This assumes that the tokens are ordered in a raster scan order, or some other order that preserves spatial locality, which is typically the case in ViT implementations. However, this reshaping step is not trivial, as it assumes that the token embeddings can be spatially rearranged into a 2D grid that corresponds to the original image patches. This assumption may not hold for all feature representations, particularly those that have undergone substantial global contextual mixing through self-attention operations.

To overlay the heatmap on the input image, the resulting heatmap is typically upsampled to the input resolution using either bilinear interpolation or nearest neighbor interpolation, and then normalized to the range [0, 1] for visualization purposes, using the following formula:
\begin{equation}
L^{(l,c)}_{\text{overlay}} = \frac{L^{(l,c)}_{\text{Grad-CAM}} - \min(L^{(l,c)}_{\text{Grad-CAM}})}{\max(L^{(l,c)}_{\text{Grad-CAM}}) - \min(L^{(l,c)}_{\text{Grad-CAM}})}\in [0,1]^{H \times W}\label{normalization}
\end{equation}

To illustrate (See Figs.~\ref{fig:vit_other_gradcam} and~\ref{fig:vit_attention_gradcam}) the practical consequences of these choices, we computed qualitative Grad-CAM examples across the feature locations defined in this taxonomy for a standard ViT-B/16~\cite{ViT} model trained on ImageNet-1k~\cite{imagenet_challenge}. Because the full set of visualizations is too large for a single page, the results are divided across two figures. The first shows $\mathcal{F}_1$, $\mathcal{F}_4$, $\mathcal{F}_5$, and $\mathcal{F}_6$ across all encoder layers, with both 

\onecolumn
\begin{landscape}

\begin{figure}[p]
    \centering
    \includegraphics[width=\linewidth,height=0.99\textheight,keepaspectratio]{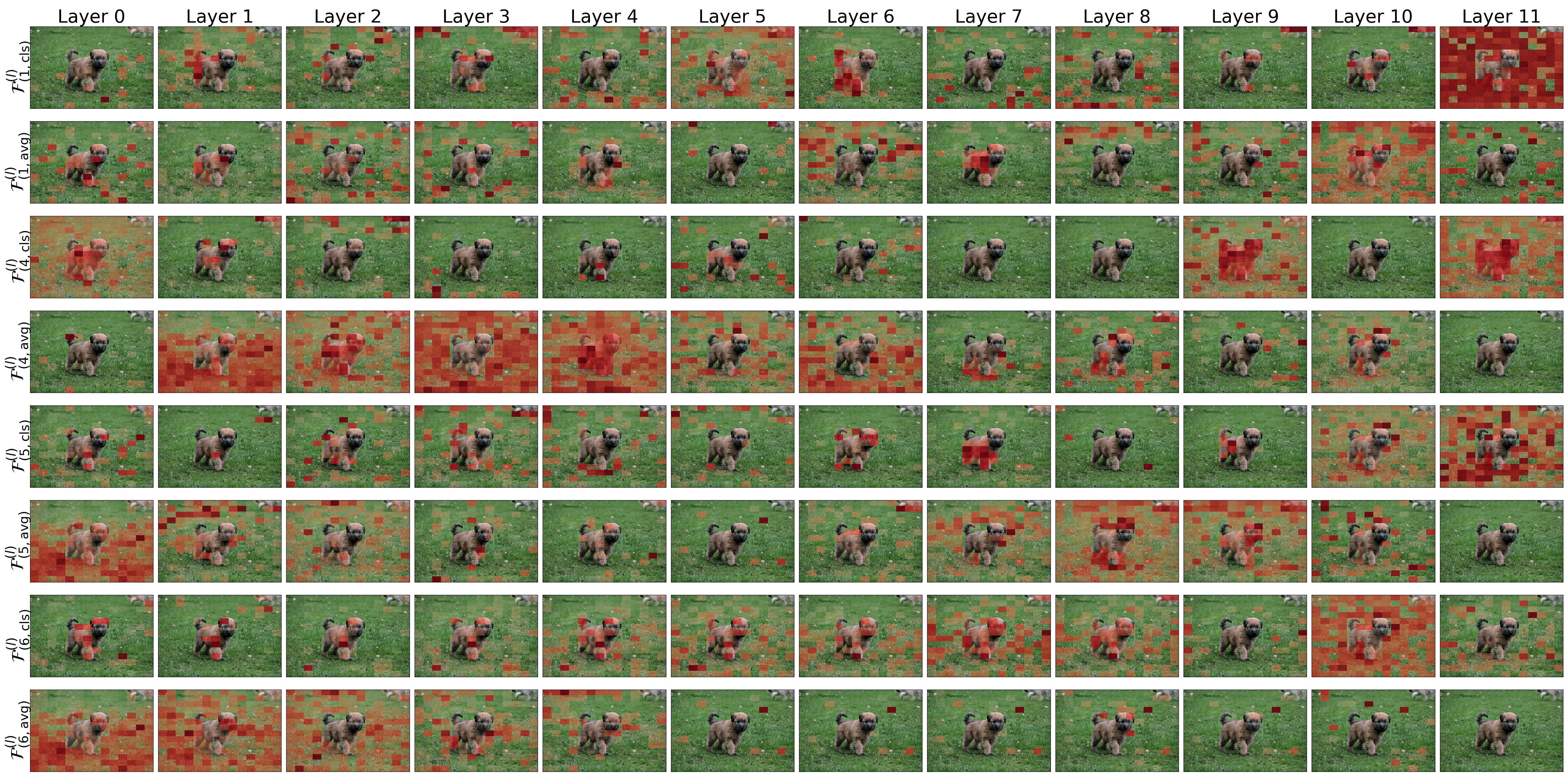}

    \caption{Grad-CAM visualizations for a standard ViT-B/16~\cite{ViT} model across feature extraction locations $\mathcal{F}_1$, $\mathcal{F}_4$, $\mathcal{F}_5$, and $\mathcal{F}_6$, comparing \texttt{[CLS]}-gradient weighting with patch-token-average gradient weighting. Rows correspond to encoder layers, and columns correspond to feature extraction locations. Differences across heatmaps show that feature location and gradient-weighting strategy can substantially change the resulting visualization.
 }
    \label{fig:vit_other_gradcam}
\end{figure}

\begin{figure}[p]
    \centering
    \includegraphics[width=\linewidth,height=0.99\textheight,keepaspectratio]{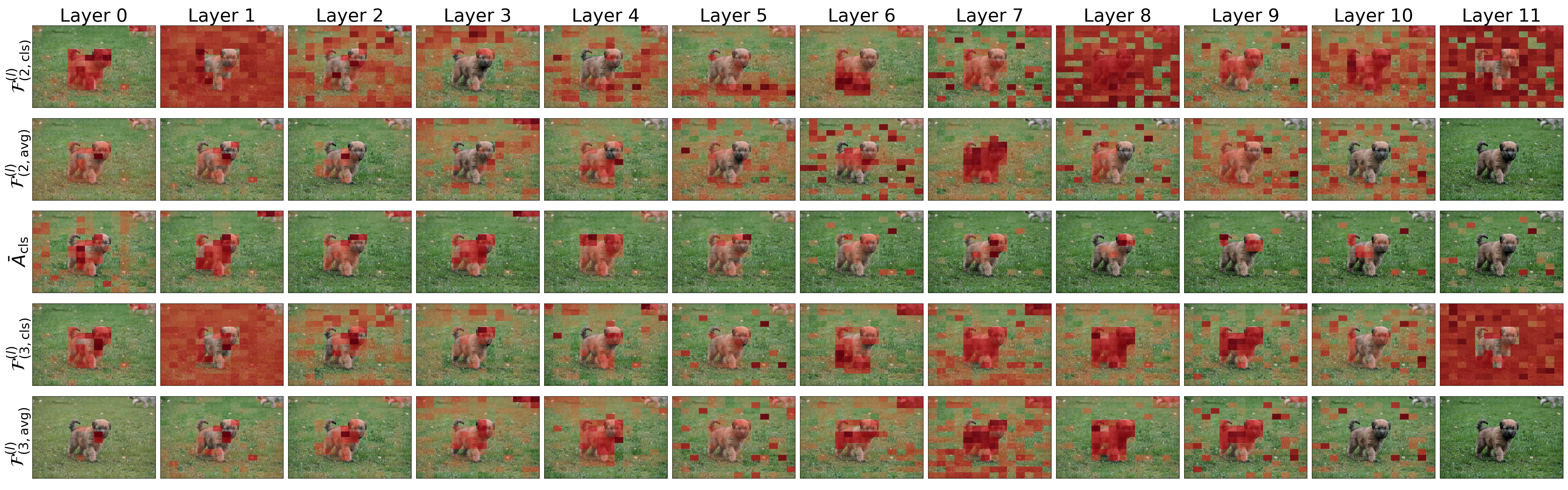}

    \caption{Grad-CAM visualizations for a standard ViT-B/16~\cite{ViT} model across feature extraction locations $\mathcal{F}_2$ and $\mathcal{F}3$, comparing \texttt{[CLS]}-gradient weighting with patch-token-average gradient weighting. The grid also includes head-averaged raw \texttt{[CLS]}-to-patch softmax attention without gradient weighting, denoted $\hat{A}_{\text{cls}}$. Rows correspond to encoder layers, and columns correspond to feature extraction locations or attention variants. Differences across heatmaps show that feature location, gradient-weighting strategy, and raw attention selection can substantially change the resulting visualization.}
    \label{fig:vit_attention_gradcam}
\end{figure}

\end{landscape}
\twocolumn

\noindent \texttt{[CLS]}-gradient weighting and patch-token-average gradient weighting. The second shows $\mathcal{F}_2$ and $\mathcal{F}_3$ under the same weighting choices, along with $\hat{A}_{\text{cls}}$, the raw softmax attention from the \texttt{[CLS]} token to patch tokens without gradient weighting.  For the target class ($y^c$), we used the model's top predicted class, which corresponded to the correct label for the input image: \textit{soft-coated wheaten terrier}.

These examples are not intended as a benchmark or as evidence that one feature location is universally best. Rather, they demonstrate that different feature locations, layers, and gradient-weighting choices can produce qualitatively different heatmaps for the same image and class. Some maps concentrate on the object or region of interest, while others highlight surrounding context or appear almost inverted relative to the object. Intermediate layers often produce more visually plausible localization than the final layer, and the final-layer token-weighted maps can become nearly blank because, in standard ViT classification, only the \texttt{[CLS]} token is passed to the classification head.

This qualitative variation exposes a major reproducibility problem. Under the taxonomy proposed here, even a standard ViT-B/16~\cite{ViT} model yields well over 100 plausible Grad-CAM-style mappings when one considers multiple encoder layers, multiple feature extraction locations, \texttt{[CLS]}-gradient versus patch-token-average weighting, and raw attention variants. If a paper reports only that it “uses Grad-CAM” on a ViT, without specifying the selected layer, feature representation, gradient target, and weighting strategy, then the reported heatmap is not a uniquely defined explanation. It is one visualization selected from a large space of possible alternatives. This creates substantial researcher degrees of freedom: authors could, intentionally or unintentionally, choose the map that appears most visually convincing while leaving readers unable to reproduce, evaluate, or even identify the actual attribution procedure. In this setting, underspecification is not a minor reporting issue; it directly affects whether the explanation can be treated as scientific evidence. These ambiguities become even more consequential in multimodal and cross-attention settings, where the attribution source may encode interactions between image tokens and language tokens rather than image-token evidence alone.

\subsubsection{Transferring to Cross-Attention and Other ViT-Based Architectures}\label{sec:cross_attention_transfer}

The above generalizations of Grad-CAM~\cite{grad_cam} for ViTs can be extended to other token-based vision architectures. For brevity, we focus on the common structure of transformer encoder blocks used in many ViT-based architectures, but the same principles apply to transformer-based vision models that rely on tokenized image representations, such as Swin Transformers~\cite{Liu_2021_ICCV}, BEiT~\cite{beit}, and the vision encoders used in vision-language models such as CLIP~\cite{clip} and BLIP~\cite{blip}. However, not all vision-language models introduce cross-attention in the same way. For example, CLIP is typically formulated as a dual-encoder model that compares separate image and text embeddings through a contrastive objective, whereas other VLMs incorporate multimodal fusion modules with cross-attention. For models that do include multimodal cross-attention, the set of usable feature representations becomes more limited when focusing specifically on the cross-attention modules. See Fig.~\ref{fig:vit_grad_cam_features} for a schematic of such a cross-attention module and the feature representations that can be extracted from it.

This limitation stems from the introduction of a query sequence and a key-value sequence that originate from different modalities. For example, in a VLM with a ViT-based vision encoder and a transformer-based language encoder, the cross-attention module typically uses the vision encoder's output tokens as the key-value sequence and the language encoder's output tokens as the query sequence. Consequently, the attention maps produced within the cross-attention module primarily encode interactions between image tokens and text tokens, rather than interactions among image tokens alone. As a result, it becomes more difficult to extract spatially localized feature representations from cross-attention modules for Grad-CAM-style visualizations, since these representations are heavily influenced by cross-modal interactions rather than purely visual features.

From the feature representations discussed in Section~\ref{sec:feature_extraction_locations}, the only representations within a cross-attention module that retain the spatial structure of the vision token sequence are (1) the context-side LayerNorm output $\mathcal{F}_{(1,X)}^{(l)}$, and (2) the attention maps $\mathcal{F}_2^{(h,l)}$. These are the only tensors whose dimensions correspond to the vision-token sequence $X^{(l)}$ (assuming the key-value sequence is derived from the vision encoder and the query sequence is derived from a non-spatial modality such as text). These feature maps are given by:
\begin{equation}
\mathcal{F}_{(1,X)}^{(l)} = \mathrm{LN}_{(1,X)}\!\left(X^{(l)}\right)
\in \mathbb{R}^{N_X \times D},\label{eq:cross_attention_context_ln}
\end{equation}
and
\begin{equation}
\mathcal{F}_{2}^{(h,l)} =
\mathrm{softmax}\!\left(
\frac{Q_h^{(l)} {K_h^{(l)}}^\top}{\sqrt{d_h}}
\right)
\in \mathbb{R}^{N_Y \times N_X},\label{eq:cross_attention_attention_map}
\end{equation}
where $X^{(l)}$ denotes the context (vision) token sequence, $\mathrm{LN}_{(1,X)}$ is the LayerNorm applied to the context sequence prior to key-value projection, $N_X$ is the number of vision tokens, $D$ is the embedding dimension, and $Q_h^{(l)} \in \mathbb{R}^{N_Y \times d_h}$ and $K_h^{(l)} \in \mathbb{R}^{N_X \times d_h}$ are the query and key matrices for attention head $h$ in block $l$.

The choice of feature representation therefore becomes clear: one may use a representation derived solely from the vision tokens, such as $\mathcal{F}_{(1,X)}^{(l)}$, or a representation that explicitly encodes interactions between the vision and language tokens, such as $\mathcal{F}_2^{(h,l)}$.

Because $\mathcal{F}_2^{(h,l)}$ retains both the query-token axis and the vision-token axis, it can also be used to construct token-specific Grad-CAM maps. Let $j \in {1,\dots,N_Y}$ denote a selected query token, such as a word or subword token in the text sequence, and let $i \in {2,\dots,N_X}$ index the spatial vision tokens, excluding the vision \texttt{[CLS]} token. Let $y^c$ denote the scalar target used for backpropagation, such as the logit for the positive image-text matching (ITM) class in an ITM model, an image-text similarity score, or the score assigned to a selected answer class.

\begin{equation}
\widetilde{\alpha}_{h,j}^{(l,c)} =
\frac{1}{N_X-1}
\sum_{i=2}^{N_X}
\frac{\partial y^c}
{\partial \left[\mathcal{F}_2^{(h,l)}\right]_{j,i}}.
\label{eq:cross_attention_token_gradients}
\end{equation}

The corresponding token-specific flat cross-attention Grad-CAM map is then given by

\begin{equation}
L^{(l,c)}_{\mathrm{Flat\text{-}GCAM},j,i} =
\mathrm{ReLU}\left(
\sum_h
\widetilde{\alpha}_{h,j}^{(l,c)}
\left[\mathcal{F}_2^{(h,l)}\right]_{j,i}
\right),
\label{eq:cross_attention_token_gcam}
\end{equation}
where $i \in {2,\dots,N_X}$. Here, $L^{(l,c)}_{\mathrm{Flat\text{-}GCAM},j,i}$ is still indexed over the flattened vision-token sequence. This formulation produces a separate flat Grad-CAM map for each query token $j$. This flattened map can then be reshaped into the original patch layout used by the model, as described in Section~\ref{sec:reshaping_problem}.

A concrete BLIP-base ITM example is provided in Supplementary Fig.~\ref{fig:blip_cross_attention_gradcam}. Using the positive ITM logit as the target $y^c$, the example computes separate cross-attention Grad-CAM maps for each query token across all cross-attention layers. The resulting per-token maps show that both word choice and layer choice can substantially affect the localization pattern.

A second variant uses gradients from the query-side \texttt{[CLS]} token to compute the head weights, while still using the selected token's cross-attention map as the feature map. In this case, the importance weights are
\begin{equation}
\widehat{\alpha}_{h}^{(l,c)} =
\frac{1}{N_X-1}
\sum_{i=2}^{N_X}
\frac{\partial y^c}
{\partial \left[\mathcal{F}_2^{(h,l)}\right]_{1,i}},
\label{eq:cross_attention_cls_gradients}
\end{equation}
where index $1$ denotes the query-side \texttt{[CLS]} token. The corresponding query-token-specific flat map is
\begin{equation}
\widehat{L}^{(l,c)}_{\mathrm{Flat\text{-}GCAM},j,i} =
\mathrm{ReLU}\left(
\sum_h
\widehat{\alpha}_{h}^{(l,c)}
\left[\mathcal{F}_2^{(h,l)}\right]_{j,i}
\right),
\label{eq:cross_attention_cls_guided_token_gcam}
\end{equation}
where $i \in {2,\dots,N_X}$.

This variant is useful when the prediction score is read from the query-side \texttt{[CLS]} representation, since gradients for non-\texttt{[CLS]} query tokens may vanish in later cross-attention layers. In this case, the \texttt{[CLS]} gradients determine the head-level importance weights, while the selected query token $j$ determines the flat cross-attention map being visualized. For visualization purposes, the resulting flat map can be reshaped into the original patch layout used by the model, as described in Section~\ref{sec:reshaping_problem}.

Every subsequent computation in the cross-attention block produces outputs indexed by the query sequence length $N_Y$ rather than the vision sequence length $N_X$. Thus, these outputs do not preserve the spatial structure of the input image and are not suitable for Grad-CAM-style visualizations. For example, the attention-output representation is given by:
\begin{equation}
\mathcal{F}_{3}^{(h,l)} =
\mathrm{softmax}\!\left(
\frac{Q_h^{(l)} {K_h^{(l)}}^\top}{\sqrt{d_h}}
\right)V_h^{(l)}
\in \mathbb{R}^{N_Y \times d_h},\label{eq:cross_attention_attention_output}
\end{equation}
where $V_h^{(l)} \in \mathbb{R}^{N_X \times d_h}$ is the value matrix for head $h$. This output cannot be reshaped using Eq.~\ref{reshaping}, as the query tokens $Y^{(l)}$ generally do not correspond to a spatial grid. The same is true for the outputs of the feedforward network and the second LayerNorm.

We also note that many ViT-based architectures do not rely on a \texttt{[CLS]} token for downstream tasks. In such cases, feature extraction and gradient aggregation may require average pooling or other token-aggregation strategies. Also, in architectures such as DEiT~\cite{deit}, both the \texttt{[CLS]} token and the distillation token (\texttt{[DIST]}) may need to be considered during feature extraction.

Our goal is not to provide an exhaustive review of all ViT-based architectures or all possible feature extraction locations, but rather to present a general framework for understanding how Grad-CAM can be applied to token-based vision models, and to highlight the importance of considering architectural characteristics when applying explainability methods such as Grad-CAM. We use this taxonomy in the following audit to analyze a bounded snapshot of papers that apply Grad-CAM or Grad-CAM-adjacent methods to ViTs and ViT-based architectures.

\section{Systematic Literature Audit Design}\label{sec:methodology}

\subsection{Research Questions}\label{sec:research_questions}
The following research questions (RQ1--RQ3) guide our systematic literature audit.
\begin{enumerate}[label=\textbf{RQ\arabic*}, ref=RQ\arabic*]
    \item How is Grad-CAM~\cite{grad_cam} being applied to ViTs in the literature? Specifically, what feature extraction locations, gradient sources, and aggregation strategies are being used, and how consistently are these choices reported across studies?
    \item How are Grad-CAM visualizations being interpreted in the context of ViTs? Are authors making claims about model behavior, feature importance, or decision-making processes based on these visualizations, and are these claims justified by the methods used?
    \item What are the common pitfalls, misinterpretations, or methodological inconsistencies in the application of Grad-CAM to ViTs, and how can these be addressed to improve the reliability and interpretability of visual explanations in transformer-based vision models?
\end{enumerate}
Each of these research questions is designed to explore different aspects of the application and interpretation of Grad-CAM in the context of ViTs, with the ultimate goal of providing a comprehensive understanding of the current state of the field and identifying areas for improvement rather than to provide a definitive answer to any one question. By systematically analyzing the literature, we aim to uncover patterns, trends, and gaps in the use of Grad-CAM for ViTs, which can inform future research and best practices in the field of explainable AI for ViTs.
\begin{figure}[tp]
    \centering
    \input{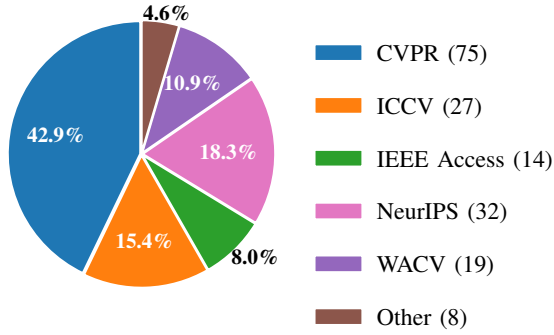}
        \caption{Distribution of surveyed literature by publication venue. This pie chart highlights the proportion of papers published in top-tier conferences (CVPR, ICCV, NeurIPS, WACV) and journals (IEEE Access), as well as other venues.}
        \label{fig:venue_pie_chart}
\end{figure}

\subsection{Paper Collection Strategy}\label{coll_strat}

Because there is no standardized search procedure for identifying papers that apply Grad-CAM~\cite{grad_cam} to ViTs, we constructed the corpus using keyword search, citation tracking, and manual screening. The initial search was conducted on April 17, 2026 using Google Scholar with the query \texttt{("Grad-CAM" OR "Gradient-weighted Class Activation Mapping") AND ("Vision Transformer" OR "ViT")}, restricted to papers from 2021 onward because ViTs were introduced in late 2020. To keep the corpus publicly accessible and focused on major computer vision, machine learning, and open-access venues, we retained papers available through the CVF Open Access repository, the NeurIPS Proceedings, or IEEE Access\footnote{See \url{https://openaccess.thecvf.com/}, \url{https://proceedings.neurips.cc/}, and \url{https://ieeeaccess.ieee.org/}.}. This filtering step reduced the corpus to approximately 550 candidate papers.

We manually screened these candidates to determine whether they actually applied Grad-CAM or a Grad-CAM-adjacent method to a ViT or ViT-based architecture. We use ``Grad-CAM'' broadly to include related attribution methods such as Grad-CAM++~\cite{grad_campp}, LayerCAM~\cite{layer_cam}, ScoreCAM~\cite{score_cam}, ScoreCAM++~\cite{score_campp}, AblationCAM~\cite{ablation_cam}, FullGrad~\cite{full_grad}, and XGradCAM~\cite{axiom_grad_cam}. We also use ``ViT'' broadly to include architectures that rely fully or partially on tokenized image representations, including Swin Transformers~\cite{Liu_2021_ICCV}, BEiT~\cite{beit}, and ViT-based vision-language models such as CLIP~\cite{clip} and BLIP~\cite{blip}. Papers using only the original CAM method~\cite{cam} were excluded because this audit focuses on Grad-CAM-style post hoc or plug-and-play attribution methods. When a surveyed paper explicitly cited or clearly relied on another work that introduced a ViT-specific Grad-CAM adaptation, that prior work was also included so that methodological dependencies could be analyzed.

\begin{figure}[tp]
\centering
\input{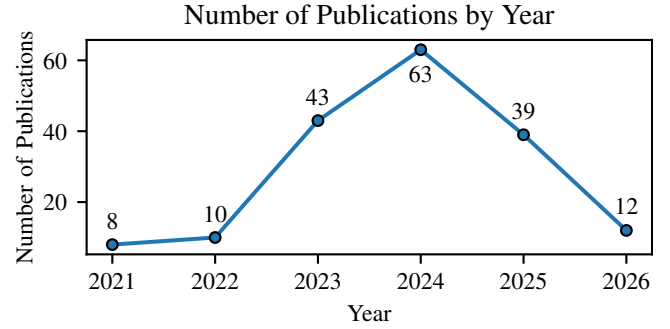}
\caption{Distribution of surveyed literature by publication year. The line plot shows the number of papers published each year that apply Grad-CAM to ViTs.}\label{fig:line_plot}
\end{figure}

The final corpus contained 175 papers that apply Grad-CAM or a Grad-CAM-adjacent method to ViTs or ViT-based architectures. We read all 175 papers and coded them for publication venue, publication year, task setting, architecture type, Grad-CAM role, citation category, methodology foundation, and reporting detail. Figures~\ref{fig:venue_pie_chart} and~\ref{fig:line_plot} summarize the corpus by publication venue and publication year, respectively. The full paper-level corpus, coding scheme, corpus construction limitations, and coding assumptions are provided in the Supplementary Materials (see~\ref{supp:limitations} and~\ref{tab:all_papers}). Because the corpus is defined by a specific search query, source filter, availability requirement, and collection date, it should be interpreted as a systematic but bounded snapshot of the literature rather than a complete census.

\subsection{Analysis Methods}\label{sec:analysis_methods}
We analyzed the final corpus using both descriptive statistics and qualitative case-study review. The quantitative analysis summarizes the distribution of papers across publication venues, publication years, task settings, architecture types, Grad-CAM roles, citation categories, and methodology foundations. These categories allow us to measure how often Grad-CAM is used on ViTs, how often authors justify or cite a ViT-specific adaptation, and whether the method is used only for visualization or as part of a broader model, pipeline, or evaluation procedure.

The qualitative analysis focuses on papers that either introduce a ViT-specific Grad-CAM adaptation or explicitly rely on a prior adaptation. For these papers, we examine the selected feature representation, gradient target, token handling, aggregation strategy, spatial reconstruction procedure, and interpretation of the resulting heatmap. The full codebook and paper-level coding details are provided in the Supplementary Materials (see~\ref{supp:coding_scheme}).

\section{Results}\label{sec:results}

\subsection{Quantitative Findings}\label{sec:quantitative_findings}

\begin{figure}[tp]
    \centering
    \resizebox{.85\columnwidth}{!}{\input{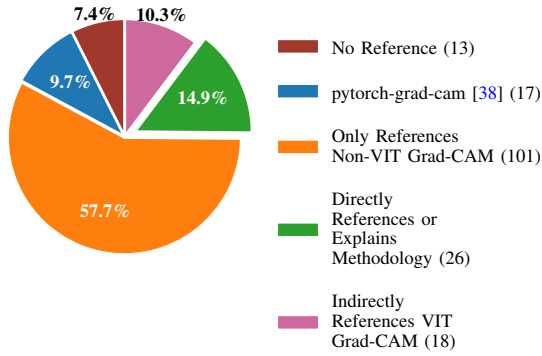}}
        \caption{Distribution of literature by Grad-CAM reference type. Categories separate explicit peer-reviewed methodology tracking from indirect citations, un-peer-reviewed code dependencies, and unreferenced baselines.}\label{fig:citation_pie_chart}
\end{figure}

Our quantitative analysis revealed that, out of the 175 papers that apply Grad-CAM to ViTs, 101 papers, or about 58\% of the total, only referenced the original Grad-CAM paper,~\cite{grad_cam}, or other Grad-CAM-adjacent methods designed for CNNs. By contrast, only 26 papers, or about 15\% of the total, either justified their use of Grad-CAM on ViTs or referenced prior work that explicitly applied Grad-CAM to ViTs (see Figure~\ref{fig:citation_pie_chart}). Eighteen papers, or about 10\%, referenced a work that justified the use of Grad-CAM on ViTs, but did not explicitly state that they used the referenced work's method. Seventeen papers, or about 10\%, only referenced the non-peer-reviewed GitHub repository \texttt{pytorch-grad-cam}~\cite{jacobgil2021pytorchgradcam}, which contains a Grad-CAM implementation for ViTs but does not provide a theoretical justification for why Grad-CAM is appropriate for transformer architectures. Finally, 13 papers, or about 7\%, did not cite any prior work on Grad-CAM and simply applied Grad-CAM to ViTs without an explicit justification, reference to prior ViT-specific work, or reference to the original Grad-CAM paper,~\cite{grad_cam}.

Taken together, these findings suggest that most surveyed papers treat Grad-CAM on ViTs as a method requiring little or no additional justification beyond the original CNN-based formulation. This is methodologically important because the phrase ``Grad-CAM on a ViT'' does not identify a unique feature representation, gradient target, token-handling procedure, or aggregation strategy. Therefore, the citation patterns in Figure~\ref{fig:citation_pie_chart} provide quantitative evidence for the central concern of this audit: ViT-based Grad-CAM visualizations are often reported without enough information to determine exactly how the heatmap was produced.

\begin{figure}[tp]
    \centering
    \resizebox{1.1\columnwidth}{!}{\input{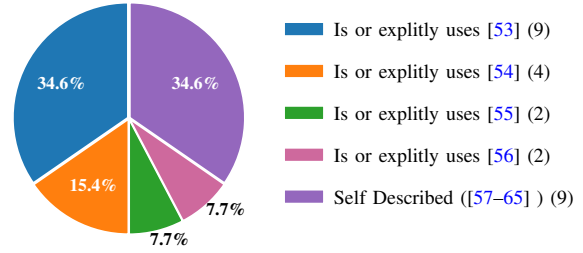}}

    \caption{Distribution of literature according to their methodology foundation if the paper Directly (or explicitly) References or Explains Methodology. Categories indicate whether a study explicitly builds upon or adopts a specific baseline framework, or introduces an independent, unadopted standalone method (Self-Described).}\label{fig:citation_pie_cites_who}
\end{figure}

Among the 26 papers that either justified their use of Grad-CAM on ViTs or explicitly cited prior work for their Grad-CAM-on-ViT methodology, 13 papers introduced or sufficiently specified a distinct method for applying Grad-CAM to ViTs. The remaining 13 papers explicitly cited one of these methods as the basis for their own Grad-CAM calculation or implementation (see Figure~\ref{fig:citation_pie_cites_who}). Among the 13 distinct Grad-CAM-on-ViT adaptations, methodological uptake was uneven. The most frequently cited method was ALBEF~\cite{NEURIPS2021_50525975}, which was cited as the Grad-CAM methodological basis by 8 surveyed papers. This was followed by Chefer et al.~\cite{Chefer_2021_CVPR}, cited by 3 papers; CLIP-ES~\cite{Lin_2023_CVPR}, cited by 1 paper; and Plug-and-Play VQA~\cite{tiong-etal-2022-plug}, cited by 1 paper. The remaining methods~\cite{NEURIPS2024_24f8dd1b,Hao_2025_ICCV,10839367,Zeng_2024_CVPR,Luo_2024_CVPR_2,Sun_2024_CVPR,Wang_AAAI_2025,Aravindan_2025_ICCV,Chen_2025_ICCV} were not cited as the Grad-CAM methodological basis by any other surveyed works. These 13 methods are analyzed qualitatively in Section~\ref{sec:qualitative_analysis}.

These results indicate that there is a small body of literature that explicitly addresses how Grad-CAM should be adapted to ViTs, but that this work has not yet become a dominant methodological reference point. Instead, most surveyed papers continue to rely on the original CNN-based Grad-CAM formulation, CNN-oriented Grad-CAM variants, or implementation-level references without giving a mathematical account of how the method transfers to transformer architectures. This pattern reinforces the central concern of this study: although Grad-CAM is widely used on ViTs, the justification for doing so is often assumed rather than demonstrated.

\begin{figure}[tp]
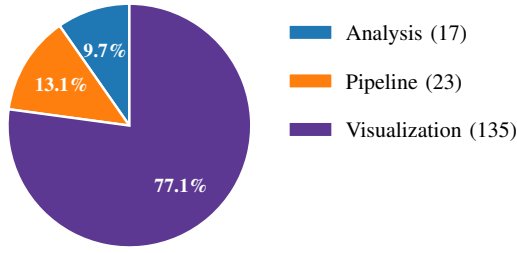

    \centering
    \resizebox{.8\columnwidth}{!}{
\begingroup%
\makeatletter%
\begin{pgfpicture}%
\pgfpathrectangle{\pgfpointorigin}{\pgfqpoint{3.034220in}{1.450000in}}%
\pgfusepath{use as bounding box, clip}%
\begin{pgfscope}%
\pgfsetbuttcap%
\pgfsetmiterjoin%
\definecolor{currentfill}{rgb}{1.000000,1.000000,1.000000}%
\pgfsetfillcolor{currentfill}%
\pgfsetlinewidth{0.000000pt}%
\definecolor{currentstroke}{rgb}{1.000000,1.000000,1.000000}%
\pgfsetstrokecolor{currentstroke}%
\pgfsetdash{}{0pt}%
\pgfpathmoveto{\pgfqpoint{0.000000in}{0.000000in}}%
\pgfpathlineto{\pgfqpoint{3.034220in}{0.000000in}}%
\pgfpathlineto{\pgfqpoint{3.034220in}{1.450000in}}%
\pgfpathlineto{\pgfqpoint{0.000000in}{1.450000in}}%
\pgfpathlineto{\pgfqpoint{0.000000in}{0.000000in}}%
\pgfpathclose%
\pgfusepath{fill}%
\end{pgfscope}%
\begin{pgfscope}%
\pgfsetbuttcap%
\pgfsetmiterjoin%
\definecolor{currentfill}{rgb}{0.121569,0.466667,0.705882}%
\pgfsetfillcolor{currentfill}%
\pgfsetlinewidth{1.003750pt}%
\definecolor{currentstroke}{rgb}{1.000000,1.000000,1.000000}%
\pgfsetstrokecolor{currentstroke}%
\pgfsetdash{}{0pt}%
\pgfpathmoveto{\pgfqpoint{0.690476in}{1.450000in}}%
\pgfpathcurveto{\pgfqpoint{0.620100in}{1.450000in}}{\pgfqpoint{0.550135in}{1.439240in}}{\pgfqpoint{0.483010in}{1.418094in}}%
\pgfpathcurveto{\pgfqpoint{0.415886in}{1.396948in}}{\pgfqpoint{0.352387in}{1.365664in}}{\pgfqpoint{0.294717in}{1.325326in}}%
\pgfpathlineto{\pgfqpoint{0.690476in}{0.759524in}}%
\pgfpathlineto{\pgfqpoint{0.690476in}{1.450000in}}%
\pgfpathclose%
\pgfusepath{stroke,fill}%
\end{pgfscope}%
\begin{pgfscope}%
\pgfsetbuttcap%
\pgfsetmiterjoin%
\definecolor{currentfill}{rgb}{1.000000,0.498039,0.054902}%
\pgfsetfillcolor{currentfill}%
\pgfsetlinewidth{1.003750pt}%
\definecolor{currentstroke}{rgb}{1.000000,1.000000,1.000000}%
\pgfsetstrokecolor{currentstroke}%
\pgfsetdash{}{0pt}%
\pgfpathmoveto{\pgfqpoint{0.294717in}{1.325326in}}%
\pgfpathcurveto{\pgfqpoint{0.216569in}{1.270664in}}{\pgfqpoint{0.150587in}{1.200400in}}{\pgfqpoint{0.100940in}{1.118975in}}%
\pgfpathcurveto{\pgfqpoint{0.051293in}{1.037549in}}{\pgfqpoint{0.019050in}{0.946713in}}{\pgfqpoint{0.006249in}{0.852209in}}%
\pgfpathlineto{\pgfqpoint{0.690476in}{0.759524in}}%
\pgfpathlineto{\pgfqpoint{0.294717in}{1.325326in}}%
\pgfpathclose%
\pgfusepath{stroke,fill}%
\end{pgfscope}%
\begin{pgfscope}%
\pgfsetbuttcap%
\pgfsetmiterjoin%
\definecolor{currentfill}{rgb}{0.360784,0.219608,0.572549}%
\pgfsetfillcolor{currentfill}%
\pgfsetlinewidth{1.003750pt}%
\definecolor{currentstroke}{rgb}{1.000000,1.000000,1.000000}%
\pgfsetstrokecolor{currentstroke}%
\pgfsetdash{}{0pt}%
\pgfpathmoveto{\pgfqpoint{0.006249in}{0.852209in}}%
\pgfpathcurveto{\pgfqpoint{-0.003128in}{0.782984in}}{\pgfqpoint{-0.001946in}{0.712735in}}{\pgfqpoint{0.009756in}{0.643865in}}%
\pgfpathcurveto{\pgfqpoint{0.021457in}{0.574995in}}{\pgfqpoint{0.043543in}{0.508297in}}{\pgfqpoint{0.075257in}{0.446054in}}%
\pgfpathcurveto{\pgfqpoint{0.106972in}{0.383811in}}{\pgfqpoint{0.147949in}{0.326740in}}{\pgfqpoint{0.196789in}{0.276792in}}%
\pgfpathcurveto{\pgfqpoint{0.245628in}{0.226844in}}{\pgfqpoint{0.301765in}{0.184597in}}{\pgfqpoint{0.363281in}{0.151494in}}%
\pgfpathcurveto{\pgfqpoint{0.424797in}{0.118390in}}{\pgfqpoint{0.490982in}{0.094814in}}{\pgfqpoint{0.559572in}{0.081570in}}%
\pgfpathcurveto{\pgfqpoint{0.628162in}{0.068326in}}{\pgfqpoint{0.698367in}{0.065568in}}{\pgfqpoint{0.767785in}{0.073389in}}%
\pgfpathcurveto{\pgfqpoint{0.837203in}{0.081211in}}{\pgfqpoint{0.905034in}{0.099522in}}{\pgfqpoint{0.968957in}{0.127697in}}%
\pgfpathcurveto{\pgfqpoint{1.032881in}{0.155871in}}{\pgfqpoint{1.092160in}{0.193584in}}{\pgfqpoint{1.144767in}{0.239546in}}%
\pgfpathcurveto{\pgfqpoint{1.197375in}{0.285508in}}{\pgfqpoint{1.242704in}{0.339188in}}{\pgfqpoint{1.279204in}{0.398751in}}%
\pgfpathcurveto{\pgfqpoint{1.315704in}{0.458314in}}{\pgfqpoint{1.342955in}{0.523073in}}{\pgfqpoint{1.360024in}{0.590813in}}%
\pgfpathcurveto{\pgfqpoint{1.377093in}{0.658552in}}{\pgfqpoint{1.383783in}{0.728492in}}{\pgfqpoint{1.379866in}{0.798239in}}%
\pgfpathcurveto{\pgfqpoint{1.375949in}{0.867986in}}{\pgfqpoint{1.361470in}{0.936737in}}{\pgfqpoint{1.336924in}{1.002140in}}%
\pgfpathcurveto{\pgfqpoint{1.312378in}{1.067542in}}{\pgfqpoint{1.278048in}{1.128843in}}{\pgfqpoint{1.235108in}{1.183945in}}%
\pgfpathcurveto{\pgfqpoint{1.192169in}{1.239047in}}{\pgfqpoint{1.141114in}{1.287314in}}{\pgfqpoint{1.083692in}{1.327097in}}%
\pgfpathcurveto{\pgfqpoint{1.026269in}{1.366879in}}{\pgfqpoint{0.963140in}{1.397718in}}{\pgfqpoint{0.896464in}{1.418558in}}%
\pgfpathcurveto{\pgfqpoint{0.829788in}{1.439399in}}{\pgfqpoint{0.760333in}{1.450000in}}{\pgfqpoint{0.690476in}{1.450000in}}%
\pgfpathlineto{\pgfqpoint{0.690476in}{0.759524in}}%
\pgfpathlineto{\pgfqpoint{0.006249in}{0.852209in}}%
\pgfpathclose%
\pgfusepath{stroke,fill}%
\end{pgfscope}%
\begin{pgfscope}%
\definecolor{textcolor}{rgb}{1.000000,1.000000,1.000000}%
\pgfsetstrokecolor{textcolor}%
\pgfsetfillcolor{textcolor}%
\pgftext[x=0.555623in,y=1.187595in,,]{\color{textcolor}{\rmfamily\fontsize{8.000000}{9.600000}\bfseries\selectfont\catcode`\^=\active\def^{\ifmmode\sp\else\^{}\fi}\catcode`\%=\active\def
\end{pgfscope}%
\begin{pgfscope}%
\definecolor{textcolor}{rgb}{1.000000,1.000000,1.000000}%
\pgfsetstrokecolor{textcolor}%
\pgfsetfillcolor{textcolor}%
\pgftext[x=0.307278in,y=0.993167in,,]{\color{textcolor}{\rmfamily\fontsize{8.000000}{9.600000}\bfseries\selectfont\catcode`\^=\active\def^{\ifmmode\sp\else\^{}\fi}\catcode`\%=\active\def
\end{pgfscope}%
\begin{pgfscope}%
\definecolor{textcolor}{rgb}{1.000000,1.000000,1.000000}%
\pgfsetstrokecolor{textcolor}%
\pgfsetfillcolor{textcolor}%
\pgftext[x=0.985765in,y=0.421538in,,]{\color{textcolor}{\rmfamily\fontsize{8.000000}{9.600000}\bfseries\selectfont\catcode`\^=\active\def^{\ifmmode\sp\else\^{}\fi}\catcode`\%=\active\def
\end{pgfscope}%
\begin{pgfscope}%
\pgfsetbuttcap%
\pgfsetmiterjoin%
\definecolor{currentfill}{rgb}{0.121569,0.466667,0.705882}%
\pgfsetfillcolor{currentfill}%
\pgfsetlinewidth{1.003750pt}%
\definecolor{currentstroke}{rgb}{1.000000,1.000000,1.000000}%
\pgfsetstrokecolor{currentstroke}%
\pgfsetdash{}{0pt}%
\pgfpathmoveto{\pgfqpoint{1.591500in}{1.243750in}}%
\pgfpathlineto{\pgfqpoint{1.841500in}{1.243750in}}%
\pgfpathlineto{\pgfqpoint{1.841500in}{1.331250in}}%
\pgfpathlineto{\pgfqpoint{1.591500in}{1.331250in}}%
\pgfpathlineto{\pgfqpoint{1.591500in}{1.243750in}}%
\pgfpathclose%
\pgfusepath{stroke,fill}%
\end{pgfscope}%
\begin{pgfscope}%
\definecolor{textcolor}{rgb}{0.000000,0.000000,0.000000}%
\pgfsetstrokecolor{textcolor}%
\pgfsetfillcolor{textcolor}%
\pgftext[x=1.916500in,y=1.243750in,left,base]{\color{textcolor}{\rmfamily\fontsize{9.000000}{10.800000}\selectfont\catcode`\^=\active\def^{\ifmmode\sp\else\^{}\fi}\catcode`\%=\active\def
\end{pgfscope}%
\begin{pgfscope}%
\pgfsetbuttcap%
\pgfsetmiterjoin%
\definecolor{currentfill}{rgb}{1.000000,0.498039,0.054902}%
\pgfsetfillcolor{currentfill}%
\pgfsetlinewidth{1.003750pt}%
\definecolor{currentstroke}{rgb}{1.000000,1.000000,1.000000}%
\pgfsetstrokecolor{currentstroke}%
\pgfsetdash{}{0pt}%
\pgfpathmoveto{\pgfqpoint{1.591500in}{0.968750in}}%
\pgfpathlineto{\pgfqpoint{1.841500in}{0.968750in}}%
\pgfpathlineto{\pgfqpoint{1.841500in}{1.056250in}}%
\pgfpathlineto{\pgfqpoint{1.591500in}{1.056250in}}%
\pgfpathlineto{\pgfqpoint{1.591500in}{0.968750in}}%
\pgfpathclose%
\pgfusepath{stroke,fill}%
\end{pgfscope}%
\begin{pgfscope}%
\definecolor{textcolor}{rgb}{0.000000,0.000000,0.000000}%
\pgfsetstrokecolor{textcolor}%
\pgfsetfillcolor{textcolor}%
\pgftext[x=1.916500in,y=0.968750in,left,base]{\color{textcolor}{\rmfamily\fontsize{9.000000}{10.800000}\selectfont\catcode`\^=\active\def^{\ifmmode\sp\else\^{}\fi}\catcode`\%=\active\def
\end{pgfscope}%
\begin{pgfscope}%
\pgfsetbuttcap%
\pgfsetmiterjoin%
\definecolor{currentfill}{rgb}{0.360784,0.219608,0.572549}%
\pgfsetfillcolor{currentfill}%
\pgfsetlinewidth{1.003750pt}%
\definecolor{currentstroke}{rgb}{1.000000,1.000000,1.000000}%
\pgfsetstrokecolor{currentstroke}%
\pgfsetdash{}{0pt}%
\pgfpathmoveto{\pgfqpoint{1.591500in}{0.693750in}}%
\pgfpathlineto{\pgfqpoint{1.841500in}{0.693750in}}%
\pgfpathlineto{\pgfqpoint{1.841500in}{0.781250in}}%
\pgfpathlineto{\pgfqpoint{1.591500in}{0.781250in}}%
\pgfpathlineto{\pgfqpoint{1.591500in}{0.693750in}}%
\pgfpathclose%
\pgfusepath{stroke,fill}%
\end{pgfscope}%
\begin{pgfscope}%
\definecolor{textcolor}{rgb}{0.000000,0.000000,0.000000}%
\pgfsetstrokecolor{textcolor}%
\pgfsetfillcolor{textcolor}%
\pgftext[x=1.916500in,y=0.693750in,left,base]{\color{textcolor}{\rmfamily\fontsize{9.000000}{10.800000}\selectfont\catcode`\^=\active\def^{\ifmmode\sp\else\^{}\fi}\catcode`\%=\active\def
\end{pgfscope}%
\end{pgfpicture}%
\makeatother%
\endgroup%
}
    \caption{Distribution of surveyed literature by Grad-CAM application type. Slices categorize whether the framework is used for core model analysis, embedded directly into a pipeline, or utilized solely for model output visualization.}\label{fig:type}
\end{figure}

In terms of the role of Grad-CAM in the surveyed literature, 135 papers, or about 77\% of the total, used Grad-CAM solely for visualization purposes. Twenty-three papers, or about 13\%, used Grad-CAM as a component of their method or pipeline, and 17 papers, or about 10\%, used Grad-CAM for quantitative analysis or evaluation (see Figure~\ref{fig:type}). This distribution shows that Grad-CAM is still primarily used as a post hoc visualization tool. However, the presence of papers that use Grad-CAM as part of a method or evaluation pipeline also shows that Grad-CAM is increasingly being treated not only as an explanatory visualization, but also as an operational component in model design, segmentation, grounding, prompt generation, and evaluation.

\begin{figure}[tp]
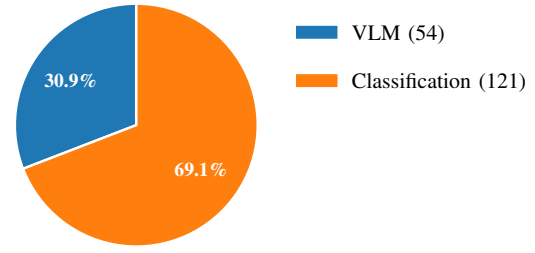

    \centering
    \resizebox{.8\columnwidth}{!}{
\begingroup%
\makeatletter%
\begin{pgfpicture}%
\pgfpathrectangle{\pgfpointorigin}{\pgfqpoint{3.045660in}{1.450000in}}%
\pgfusepath{use as bounding box, clip}%
\begin{pgfscope}%
\pgfsetbuttcap%
\pgfsetmiterjoin%
\definecolor{currentfill}{rgb}{1.000000,1.000000,1.000000}%
\pgfsetfillcolor{currentfill}%
\pgfsetlinewidth{0.000000pt}%
\definecolor{currentstroke}{rgb}{1.000000,1.000000,1.000000}%
\pgfsetstrokecolor{currentstroke}%
\pgfsetdash{}{0pt}%
\pgfpathmoveto{\pgfqpoint{0.000000in}{0.000000in}}%
\pgfpathlineto{\pgfqpoint{3.045660in}{0.000000in}}%
\pgfpathlineto{\pgfqpoint{3.045660in}{1.450000in}}%
\pgfpathlineto{\pgfqpoint{0.000000in}{1.450000in}}%
\pgfpathlineto{\pgfqpoint{0.000000in}{0.000000in}}%
\pgfpathclose%
\pgfusepath{fill}%
\end{pgfscope}%
\begin{pgfscope}%
\pgfsetbuttcap%
\pgfsetmiterjoin%
\definecolor{currentfill}{rgb}{0.121569,0.466667,0.705882}%
\pgfsetfillcolor{currentfill}%
\pgfsetlinewidth{1.003750pt}%
\definecolor{currentstroke}{rgb}{1.000000,1.000000,1.000000}%
\pgfsetstrokecolor{currentstroke}%
\pgfsetdash{}{0pt}%
\pgfpathmoveto{\pgfqpoint{0.690476in}{1.450000in}}%
\pgfpathcurveto{\pgfqpoint{0.578374in}{1.450000in}}{\pgfqpoint{0.467941in}{1.422700in}}{\pgfqpoint{0.368752in}{1.370466in}}%
\pgfpathcurveto{\pgfqpoint{0.269563in}{1.318233in}}{\pgfqpoint{0.184570in}{1.242621in}}{\pgfqpoint{0.121144in}{1.150188in}}%
\pgfpathcurveto{\pgfqpoint{0.057718in}{1.057755in}}{\pgfqpoint{0.017747in}{0.951251in}}{\pgfqpoint{0.004695in}{0.839911in}}%
\pgfpathcurveto{\pgfqpoint{-0.008356in}{0.728572in}}{\pgfqpoint{0.005902in}{0.615711in}}{\pgfqpoint{0.046232in}{0.511115in}}%
\pgfpathlineto{\pgfqpoint{0.690476in}{0.759524in}}%
\pgfpathlineto{\pgfqpoint{0.690476in}{1.450000in}}%
\pgfpathclose%
\pgfusepath{stroke,fill}%
\end{pgfscope}%
\begin{pgfscope}%
\pgfsetbuttcap%
\pgfsetmiterjoin%
\definecolor{currentfill}{rgb}{1.000000,0.498039,0.054902}%
\pgfsetfillcolor{currentfill}%
\pgfsetlinewidth{1.003750pt}%
\definecolor{currentstroke}{rgb}{1.000000,1.000000,1.000000}%
\pgfsetstrokecolor{currentstroke}%
\pgfsetdash{}{0pt}%
\pgfpathmoveto{\pgfqpoint{0.046232in}{0.511115in}}%
\pgfpathcurveto{\pgfqpoint{0.091472in}{0.393785in}}{\pgfqpoint{0.167917in}{0.291013in}}{\pgfqpoint{0.267279in}{0.213941in}}%
\pgfpathcurveto{\pgfqpoint{0.366640in}{0.136868in}}{\pgfqpoint{0.485194in}{0.088383in}}{\pgfqpoint{0.610089in}{0.073743in}}%
\pgfpathcurveto{\pgfqpoint{0.734983in}{0.059103in}}{\pgfqpoint{0.861536in}{0.078856in}}{\pgfqpoint{0.976029in}{0.130861in}}%
\pgfpathcurveto{\pgfqpoint{1.090521in}{0.182866in}}{\pgfqpoint{1.188660in}{0.265173in}}{\pgfqpoint{1.259808in}{0.368860in}}%
\pgfpathcurveto{\pgfqpoint{1.330956in}{0.472547in}}{\pgfqpoint{1.372445in}{0.593726in}}{\pgfqpoint{1.379778in}{0.719262in}}%
\pgfpathcurveto{\pgfqpoint{1.387110in}{0.844797in}}{\pgfqpoint{1.360011in}{0.969983in}}{\pgfqpoint{1.301419in}{1.081248in}}%
\pgfpathcurveto{\pgfqpoint{1.242826in}{1.192513in}}{\pgfqpoint{1.154937in}{1.285686in}}{\pgfqpoint{1.047277in}{1.350667in}}%
\pgfpathcurveto{\pgfqpoint{0.939618in}{1.415648in}}{\pgfqpoint{0.816226in}{1.450000in}}{\pgfqpoint{0.690476in}{1.450000in}}%
\pgfpathlineto{\pgfqpoint{0.690476in}{0.759524in}}%
\pgfpathlineto{\pgfqpoint{0.046232in}{0.511115in}}%
\pgfpathclose%
\pgfusepath{stroke,fill}%
\end{pgfscope}%
\begin{pgfscope}%
\definecolor{textcolor}{rgb}{1.000000,1.000000,1.000000}%
\pgfsetstrokecolor{textcolor}%
\pgfsetfillcolor{textcolor}%
\pgftext[x=0.320410in,y=1.013456in,,]{\color{textcolor}{\rmfamily\fontsize{8.000000}{9.600000}\bfseries\selectfont\catcode`\^=\active\def^{\ifmmode\sp\else\^{}\fi}\catcode`\%=\active\def
\end{pgfscope}%
\begin{pgfscope}%
\definecolor{textcolor}{rgb}{1.000000,1.000000,1.000000}%
\pgfsetstrokecolor{textcolor}%
\pgfsetfillcolor{textcolor}%
\pgftext[x=1.060542in,y=0.505592in,,]{\color{textcolor}{\rmfamily\fontsize{8.000000}{9.600000}\bfseries\selectfont\catcode`\^=\active\def^{\ifmmode\sp\else\^{}\fi}\catcode`\%=\active\def
\end{pgfscope}%
\begin{pgfscope}%
\pgfsetbuttcap%
\pgfsetmiterjoin%
\definecolor{currentfill}{rgb}{0.121569,0.466667,0.705882}%
\pgfsetfillcolor{currentfill}%
\pgfsetlinewidth{1.003750pt}%
\definecolor{currentstroke}{rgb}{1.000000,1.000000,1.000000}%
\pgfsetstrokecolor{currentstroke}%
\pgfsetdash{}{0pt}%
\pgfpathmoveto{\pgfqpoint{1.591500in}{1.243750in}}%
\pgfpathlineto{\pgfqpoint{1.841500in}{1.243750in}}%
\pgfpathlineto{\pgfqpoint{1.841500in}{1.331250in}}%
\pgfpathlineto{\pgfqpoint{1.591500in}{1.331250in}}%
\pgfpathlineto{\pgfqpoint{1.591500in}{1.243750in}}%
\pgfpathclose%
\pgfusepath{stroke,fill}%
\end{pgfscope}%
\begin{pgfscope}%
\definecolor{textcolor}{rgb}{0.000000,0.000000,0.000000}%
\pgfsetstrokecolor{textcolor}%
\pgfsetfillcolor{textcolor}%
\pgftext[x=1.916500in,y=1.243750in,left,base]{\color{textcolor}{\rmfamily\fontsize{9.000000}{10.800000}\selectfont\catcode`\^=\active\def^{\ifmmode\sp\else\^{}\fi}\catcode`\%=\active\def
\end{pgfscope}%
\begin{pgfscope}%
\pgfsetbuttcap%
\pgfsetmiterjoin%
\definecolor{currentfill}{rgb}{1.000000,0.498039,0.054902}%
\pgfsetfillcolor{currentfill}%
\pgfsetlinewidth{1.003750pt}%
\definecolor{currentstroke}{rgb}{1.000000,1.000000,1.000000}%
\pgfsetstrokecolor{currentstroke}%
\pgfsetdash{}{0pt}%
\pgfpathmoveto{\pgfqpoint{1.591500in}{0.968750in}}%
\pgfpathlineto{\pgfqpoint{1.841500in}{0.968750in}}%
\pgfpathlineto{\pgfqpoint{1.841500in}{1.056250in}}%
\pgfpathlineto{\pgfqpoint{1.591500in}{1.056250in}}%
\pgfpathlineto{\pgfqpoint{1.591500in}{0.968750in}}%
\pgfpathclose%
\pgfusepath{stroke,fill}%
\end{pgfscope}%
\begin{pgfscope}%
\definecolor{textcolor}{rgb}{0.000000,0.000000,0.000000}%
\pgfsetstrokecolor{textcolor}%
\pgfsetfillcolor{textcolor}%
\pgftext[x=1.916500in,y=0.968750in,left,base]{\color{textcolor}{\rmfamily\fontsize{9.000000}{10.800000}\selectfont\catcode`\^=\active\def^{\ifmmode\sp\else\^{}\fi}\catcode`\%=\active\def
\end{pgfscope}%
\end{pgfpicture}%
\makeatother%
\endgroup%
}
    \caption{Distribution of surveyed literature across primary task paradigms. Slices compare the application of the framework between traditional vision classification tasks and modern VLMs.}\label{fig:type_2}
\end{figure}

In terms of task type, 121 papers, or about 69\% of the total, applied Grad-CAM to traditional vision classification or segmentation tasks, while 54 papers, or about 31\%, applied Grad-CAM to VLMs (see Figure~\ref{fig:type_2}). This distinction is important because VLMs introduce additional ambiguity into Grad-CAM adaptation. In single-modal ViTs, the main question is usually which visual token representation should be treated as the feature map. In VLMs, however, Grad-CAM may be applied to image-encoder token activations, visual self-attention maps, multimodal cross-attention maps, or text-conditioned similarity scores. The VLM subset therefore demonstrates that the adaptation of Grad-CAM to ViTs is not a single methodological problem, but a family of related problems depending on the architecture and target objective.

Finally, 165 papers, or about 94\% of the total, used Grad-CAM on purely token-based ViTs, while only 10 papers, or about 6\%, used Grad-CAM on hybrid architectures that combine convolutional and transformer components. This finding further emphasizes the need for ViT-specific analysis. In hybrid architectures, convolutional components may still provide feature maps that resemble the original setting of Grad-CAM. In purely token-based ViTs, by contrast, the spatial feature map assumed by Grad-CAM must be reconstructed or reinterpreted through token reshaping, attention structure, or another transformer-specific representation.

Surveyed papers applying Grad-CAM to ViTs were concentrated in 2024, with smaller numbers appearing in 2021, 2022, 2023, 2025, and 2026 (see Figure~\ref{fig:line_plot}). The apparent decline after 2024 should be interpreted cautiously because the study was conducted in mid-2026 and the 2026 count is incomplete. It may also reflect a broader shift toward transformer-specific explanation methods, such as Attention Rollout~\cite{attention_rollout}, Transformer Attribution~\cite{Chefer_2021_CVPR}, and other attention-based attribution techniques. Overall, this timing supports the central finding of this audit: Grad-CAM was rapidly adopted for transformer architectures before reporting conventions and mathematical justifications for ViT-specific adaptations had fully stabilized.

\subsection{Qualitative Analysis of ViT Grad-CAM Adaptation Patterns}\label{sec:qualitative_analysis}

This section qualitatively examines the distinct technical patterns that appear among ViT-specific Grad-CAM adaptations. We focus on 13 methods that either introduce a ViT-specific Grad-CAM adaptation or provide enough methodological detail to support a paper-level taxonomy placement. Papers that only cite Grad-CAM, briefly mention a related method, or present heatmaps without sufficient implementation detail are therefore included in the broader quantitative audit but not treated as distinct technical adaptations in this qualitative analysis. Rather than presenting the analysis paper by paper, we organize the findings around the three research questions: how Grad-CAM is technically adapted to ViTs, how the resulting visualizations are interpreted, and what methodological ambiguities remain. Detailed paper-level mathematical mappings for the 13 methods examined in depth are provided in the Supplementary Materials (see\ref{supp:paper_level_mapping}).

\subsubsection{RQ1: Technical Patterns in ViT Grad-CAM Adaptation}\label{sec:rq1_technical_patterns}

\paragraph{Feature extraction locations}

Across the distinct ViT-specific Grad-CAM adaptations examined in depth, the most important technical difference is the choice of feature representation. In CNN Grad-CAM, the target feature map is usually a convolutional activation tensor with a clear spatial and channel structure. In ViTs, however, there is no single equivalent tensor. The reviewed adaptations therefore select from several transformer representations, including token embeddings, layer-normalized token features, MLP outputs, residual-stream activations, self-attention maps, cross-attention maps, and already spatialized attention maps.

A first group of methods adapts Grad-CAM to token-feature representations~\cite{Hao_2025_ICCV,10839367,NEURIPS2024_24f8dd1b,Sun_2024_CVPR,Lin_2023_CVPR,Aravindan_2025_ICCV}. In many token-feature methods, the patch-token axis provides the spatial structure, while the embedding or feature dimension is treated as analogous to the CNN channel axis. Other token-feature adaptations instead compute token-level weights or directly combine gradients and intermediate features. These methods are closest to the token-activation branch of the taxonomy, including representations such as $\mathcal{F}_1^{(l)}$, $\mathcal{F}_4^{(l)}$, $\mathcal{F}_5^{(l)}$, and $\mathcal{F}_6^{(l)}$. The resulting heatmaps are therefore best interpreted as patch-token or feature-activation attribution, not as attribution over a native convolutional feature map. Compared with attention-map adaptations, this branch remains closer to the structure of classical Grad-CAM, but still requires additional choices about how to remove non-spatial tokens, reshape patch tokens into an image grid, and identify the exact transformer block location used for attribution.

A second group of methods adapts Grad-CAM to attention-based representations~\cite{Chefer_2021_CVPR,tiong-etal-2022-plug,NEURIPS2021_50525975,Chen_2025_ICCV,Wang_AAAI_2025,Luo_2024_CVPR_2,Zeng_2024_CVPR}. These methods are closest to the $\mathcal{F}_2^{(h,l)}$ branch of the taxonomy because they treat self-attention or cross-attention maps as the attribution-bearing object. In single-modal ViTs, this often means using \texttt{[CLS]}-conditioned attention from the classification token to patch tokens. In VLMs, it often means using cross-attention maps that connect text tokens with image patches. These attention-map adaptations differ from token-feature methods because the attribution is computed over attention entries rather than over embedding channels. As a result, the selected query token, key-token set, attention head, layer, and head-aggregation strategy become central parts of the Grad-CAM adaptation.

A third pattern is that several papers use intermediate activations or attention maps without specifying the exact extraction location~\cite{10839367,NEURIPS2024_24f8dd1b,Luo_2024_CVPR_2,Zeng_2024_CVPR,Aravindan_2025_ICCV}. In these cases, the method can often be placed broadly within the taxonomy, but not assigned precisely to a single feature representation.

\paragraph{Gradient targets}
The reviewed adaptations use a relatively small set of gradient targets, but these targets still matter for reproducibility. In classification settings, Grad-CAM is often differentiated with respect to a class score, true-label score, or task-specific output~\cite{Hao_2025_ICCV,10839367,Chefer_2021_CVPR}. In segmentation and VLM settings, the target may instead be a segmentation-related score, ITM score, matching objective, task-specific loss, or softmax-normalized text/class score~\cite{Lin_2023_CVPR,tiong-etal-2022-plug,NEURIPS2021_50525975,Chen_2025_ICCV,Wang_AAAI_2025,Luo_2024_CVPR_2,Zeng_2024_CVPR,Sun_2024_CVPR,Aravindan_2025_ICCV}. These choices are closely tied to the task being explained: classification methods generally ask which image tokens support a class decision, while VLM methods ask which image regions support an image-text match, caption token, noun phrase, or grounding objective.

\paragraph{Aggregation strategies}
Aggregation strategies varied according to whether the selected representation was a token-feature tensor or an attention map. Token-feature methods usually preserve the general structure of classical Grad-CAM by averaging gradients over spatial, patch-token, or spatiotemporal locations to obtain feature-dimension weights, then forming a weighted activation map~\cite{Hao_2025_ICCV,Sun_2024_CVPR,Lin_2023_CVPR,Aravindan_2025_ICCV}. Attention-based methods often use a different strategy: rather than computing channel-wise weights, they apply gradients directly to attention entries and then aggregate across attention heads, text tokens, or patch tokens, while also depending on the selected layer~\cite{tiong-etal-2022-plug,NEURIPS2021_50525975,Chen_2025_ICCV,Wang_AAAI_2025,Luo_2024_CVPR_2,Zeng_2024_CVPR}. This distinction is important because averaging over embedding dimensions, attention heads, query tokens, or spatial tokens does not produce equivalent heatmaps.

Across the reviewed adaptations, aggregation was one of the most common places where implementation details became underspecified. Some papers identify the selected feature representation but do not clearly state whether heads are averaged before or after gradient weighting, whether text-token relevance is computed separately or pooled, or how patch-token scores are reshaped into the final image grid~\cite{Chefer_2021_CVPR,10839367,NEURIPS2024_24f8dd1b,Luo_2024_CVPR_2,Zeng_2024_CVPR}. Thus, for RQ1, aggregation strategy is a key reporting requirement because it determines how transformer-level relevance is reduced into a two-dimensional visualization.

\subsubsection{RQ2: Interpretation of Grad-CAM Visualizations in ViTs}
\label{sec:rq2_interpretation_patterns}

\paragraph{Heatmaps as localization and token-importance evidence}

In classification and segmentation settings, Grad-CAM visualizations are often interpreted as evidence that a ViT focuses on task-relevant image regions~\cite{Chefer_2021_CVPR,10839367,Lin_2023_CVPR}. This interpretation is most direct when the selected representation preserves patch-token identity and the final heatmap can be mapped back to the image grid. However, the technical meaning of these maps varies across adaptations. Some methods produce patch-token or spatiotemporal-token importance scores, while others produce \texttt{[CLS]}-conditioned attention maps or activation-based feature maps~\cite{Hao_2025_ICCV,Chefer_2021_CVPR,10839367,Sun_2024_CVPR,Lin_2023_CVPR,Aravindan_2025_ICCV}. These outputs can all be visualized as heatmaps, but they do not necessarily explain the same object. A token-activation map indicates which patch features contribute to the target score, whereas a \texttt{[CLS]}-conditioned attention map reflects how the classification token is connected to image patches. Therefore, when authors use Grad-CAM visualizations as localization evidence, the strength of that interpretation depends on whether the selected feature representation, token handling, and reshaping procedure support a spatial claim.

\paragraph{Heatmaps as visual grounding}

In VLM settings, Grad-CAM visualizations are often interpreted as evidence of visual grounding rather than as ordinary image-only localization~\cite{tiong-etal-2022-plug,NEURIPS2021_50525975,Wang_AAAI_2025,Luo_2024_CVPR_2,Zeng_2024_CVPR}. These maps are usually conditioned on language through a caption, question, noun phrase, referring expression, ITM objective, or image-text similarity score. As a result, the heatmap does not simply indicate which image regions are visually important in isolation; it indicates which image regions contribute to a particular image-text relationship. This distinction is especially important for cross-attention methods, where relevance is computed over interactions between text tokens and image patches~\cite{tiong-etal-2022-plug,NEURIPS2021_50525975,Chen_2025_ICCV,Wang_AAAI_2025}. A grounding map for the word ``dog,'' for example, has a different interpretation than a class-level map for an image classifier or a generic attention map from a visual encoder. Therefore, when Grad-CAM is used to support claims about grounding, authors should specify the language-conditioned target, the query tokens or text units involved, and whether relevance is aggregated across words, heads, or layers.

\paragraph{Heatmaps as pipeline components}

Several reviewed methods use Grad-CAM-derived maps as components within a larger pipeline rather than as post hoc visual explanations. In these cases, the heatmap may guide patch masking, generate an initial localization prompt, support mask proposal generation, or contribute to iterative segmentation refinement~\cite{NEURIPS2024_24f8dd1b,Chen_2025_ICCV,Sun_2024_CVPR,Wang_AAAI_2025}. This use changes the evidentiary role of the visualization. When a Grad-CAM map is shown only as a qualitative explanation, ambiguity affects how readers interpret the figure. When the map is used inside a pipeline, ambiguity can also affect downstream predictions, masks, prompts, or selected tokens. For this reason, pipeline-based uses require especially clear reporting of the selected feature representation, gradient target, aggregation strategy, and spatial reconstruction step.

\subsubsection{RQ3: Methodological Pitfalls and Sources of Ambiguity}
\label{sec:rq3_pitfalls}

\paragraph{Underspecified attribution tensor and token handling}

A common source of ambiguity was incomplete reporting of the attribution tensor and token handling procedure. Several methods could be placed broadly within the taxonomy, but not assigned to a precise feature location because the paper did not fully specify whether the selected representation came from token embeddings, MLP outputs, residual-stream activations, attention maps, or another intermediate tensor~\cite{10839367,NEURIPS2024_24f8dd1b,Luo_2024_CVPR_2,Zeng_2024_CVPR,Aravindan_2025_ICCV}. In some cases, the treatment of non-spatial tokens was also unclear, including whether the \texttt{[CLS]} token was removed, used as the query token, or included before reshaping patch-token scores into an image grid~\cite{10839367,NEURIPS2024_24f8dd1b,Sun_2024_CVPR}. These ambiguities were especially important for methods that described the Grad-CAM input only as a ``feature map'', ``activation,'' or ``attention map'' without identifying the exact transformer representation.

\paragraph{Ambiguous aggregation and gradient specification}

Another recurring ambiguity was incomplete reporting of how gradients and transformer dimensions were aggregated. Some methods clearly described whether relevance was averaged over spatial tokens, attention heads, text tokens, or feature dimensions, but others left these reduction steps implicit~\cite{Chefer_2021_CVPR,10839367,NEURIPS2024_24f8dd1b,Luo_2024_CVPR_2,Zeng_2024_CVPR}. This was especially common in attention-based adaptations, where the final heatmap may depend on whether heads are averaged before or after gradient weighting, whether token-specific maps are preserved or pooled, and which layer is selected for attribution~\cite{Chefer_2021_CVPR,tiong-etal-2022-plug,NEURIPS2021_50525975,Luo_2024_CVPR_2,Zeng_2024_CVPR}. In a smaller number of cases, the scalar gradient target was broad or underspecified, including whether gradients were taken with respect to an image-text similarity score, ITM objective, loss, or another task-specific output~\cite{Zeng_2024_CVPR,Wang_AAAI_2025}.

\paragraph{Heuristic attention-map adaptations of Grad-CAM}

A third recurring issue was the unclear boundary between attention visualization and Grad-CAM. Several attention-based adaptations described their method as Grad-CAM or Grad-CAM-style attribution, but computed relevance by applying gradients directly to attention entries rather than by averaging gradients into channel-wise weights and forming a weighted activation map~\cite{tiong-etal-2022-plug,Chen_2025_ICCV,Wang_AAAI_2025,Luo_2024_CVPR_2,Zeng_2024_CVPR}. A related ambiguity appears in \cite{Chefer_2021_CVPR}, where the authors state that ``the best way we found to apply GradCAM was to treat the last attention layer's \texttt{[CLS]} token as the designated feature map,'' while excluding the \texttt{[CLS]} token itself. This frames the adaptation as a heuristic implementation choice rather than as a formally justified transformer analogue of CNN Grad-CAM. The issue becomes more consequential when such a heuristic implementation is used as a comparative baseline, because the comparison is then made against one particular operationalization of ``Grad-CAM'' rather than against a uniquely defined ViT Grad-CAM method. These attention-based variants are related to Grad-CAM because they combine activations with gradient information, but they differ from raw attention visualization as well as from classical channel-weighted Grad-CAM.

\paragraph{Citation without methodological inheritance}

A final ambiguity involved the relationship between citation and methodological use. Several papers cited prior Grad-CAM or ViT-specific Grad-CAM work without clearly indicating whether the cited adaptation was actually implemented. In other cases, a method was mentioned in related work but was not explicitly adopted as the basis for the reported visualization, as recorded in the supplementary coding table and subsequent individual paper-level analysis (see~\ref{tab:all_papers} and~\ref{supp:paper_level_mapping}). This distinction matters for the audit because citation alone does not establish methodological inheritance. Therefore, papers were treated as relying on a prior adaptation only when the text clearly indicated that the method was used, followed, or implemented, rather than merely cited or mentioned.

\section{Discussion}\label{sec:discussion}

In this section, we discuss the implications of our findings, the challenges in adapting Grad-CAM to ViTs, and the broader impact on the research community.

\subsection{Why This Taxonomy Matters}\label{sec:taxonomy_matters}

The importance of this taxonomy is not that Grad-CAM is impossible to apply to ViTs, but that its application is not uniquely defined by the term ``Grad-CAM'' alone. In CNNs, the activation tensor used by Grad-CAM has an explicit spatial and channel structure, so the feature map, gradient pooling operation, and final heatmap have a relatively direct interpretation. In ViTs, the corresponding object must be chosen from several possible transformer representations, including patch-token embeddings, attention matrices, attention outputs, residual-stream activations, MLP outputs, and cross-attention maps in multimodal models. Each of these choices changes what the resulting visualization represents. A heatmap derived from token embeddings is not equivalent to one derived from attention coefficients, and a cross-attention map in a VLM has a different interpretation than a visual self-attention map in a single-modal ViT.

This distinction matters because the phrase ``Grad-CAM on a ViT'' can hide substantial methodological variation. Two papers may use the same explanation label while computing attribution over entirely different tensors, layers, gradient targets, and aggregation axes. One implementation may treat the embedding dimension as the analogue of CNN channels, while another may treat attention heads as channels, multiply attention scores directly by gradients, or average relevance across text tokens in a multimodal encoder. These choices are not minor implementation details. They can change whether the heatmap localizes the object of interest, emphasizes surrounding context, becomes diffuse, or even becomes nearly blank in layers where gradients vanish for non-\texttt{[CLS]} tokens.

This variation has direct consequences for reproducibility. In this audit, we examined 175 papers that applied Grad-CAM or Grad-CAM-adjacent methods to ViTs or ViT-based architectures, yet only 26 explicitly justified the adaptation or cited prior work that did so. Among these, we identified 13 unique peer-reviewed adaptations. This suggests that the field has not converged on a stable convention for what Grad-CAM should mean in transformer-based vision models. If authors do not report the selected feature representation, gradient target, token handling, layer choice, head aggregation, and spatial reconstruction procedure, later researchers cannot reliably reproduce the visualization or determine whether differences across papers reflect model behavior or attribution implementation choices.

The problem becomes especially important when Grad-CAM visualizations are used as evidence rather than illustration. In domains such as medical image analysis, remote sensing, or safety-critical visual recognition, heatmaps may be used to support claims about whether a model attends to clinically, physically, or semantically meaningful regions. If the attribution procedure is underspecified, the heatmap can appear persuasive while remaining difficult to interpret scientifically. The concern is therefore not only visual quality, but evidentiary status: an underspecified heatmap cannot support strong claims about model reasoning, trustworthiness, or grounding because readers cannot determine exactly how the explanation was produced.

For this reason, the taxonomy proposed in this paper is descriptive rather than prescriptive. It does not identify a single correct Grad-CAM adaptation for ViTs, nor does it claim that one feature location is universally more faithful than another. Instead, it provides terminology for making the relevant methodological choices explicit. By distinguishing token-feature methods, attention-map methods, residual and MLP feature choices, head-wise aggregation strategies, and cross-attention variants in VLMs, the taxonomy helps authors describe what they did and helps reviewers evaluate whether the reported visualization is sufficiently specified. Making these assumptions visible is necessary for treating Grad-CAM visualizations on ViTs as reproducible scientific evidence rather than as loosely defined qualitative figures.

\subsection{What Would a Proper Adaptation Require?}\label{sec:proper_adaptation}
A proper adaptation of Grad-CAM to ViTs does not necessarily require a new explanation method, but it does require more than applying the original CNN-based formula to a transformer and reporting the resulting heatmap. Because ViTs do not contain convolutional feature maps in the same sense as CNNs, authors must explicitly define which transformer representation is being treated as the Grad-CAM activation map and why that representation supports spatial attribution.

First, the feature representation used for attribution must be identified. In ViTs, there is no single equivalent to the convolutional activation tensor used in CNN Grad-CAM. The attribution source may be a patch-token embedding, a layer-normalized token representation, an attention matrix, an attention output, an MLP activation, a residual-stream representation, or a cross-attention map in a multimodal model. Therefore, a paper should specify whether the selected representation corresponds to $\mathcal{F}_1^{(l)}$, $\mathcal{F}_2^{(h,l)}$, $\mathcal{F}_3^{(h,l)}$, $\mathcal{F}_4^{(l)}$, $\mathcal{F}_5^{(l)}$, $\mathcal{F}_6^{(l)}$, or another clearly defined tensor. Without this information, the statement that a paper ``uses Grad-CAM'' on a ViT is mathematically incomplete.

Second, the adaptation must explain how the selected representation maps back to image space. If the method uses patch-token embeddings, non-spatial tokens such as \texttt{[CLS]} must be removed or otherwise handled before reshaping the remaining tokens into a spatial grid. If the method uses attention maps, authors must specify which query-token and key-token relationships are visualized. A CLS-to-patch self-attention map, a text-to-image cross-attention map, and a full patch-to-patch attention matrix have different meanings and should not be treated as interchangeable spatial heatmaps.

Third, the gradient target must be explicitly defined. In CNN classification, Grad-CAM is usually computed with respect to a class logit $y^c$. In ViTs and VLMs, the target may instead be a class score, a softmax-normalized probability, an image-text similarity score, an ITM score, a segmentation objective, or another task-specific scalar. Since different targets produce different gradients, a reproducible adaptation must state which scalar objective is differentiated and whether the gradient is taken before or after softmax or other normalization operations.

Fourth, the aggregation procedure must be specified. Classical Grad-CAM averages gradients over spatial locations to obtain channel-wise weights, but ViT adaptations may aggregate over patch tokens, embedding dimensions, attention heads, query tokens, text tokens, layers, or cross-attention entries. Token-feature methods may treat the embedding dimension as the channel axis, while attention-map methods may multiply attention scores directly by gradients and then aggregate over heads or tokens. These choices are not equivalent, so a mathematically complete adaptation should state which dimensions are averaged, summed, selected, or discarded.

Fifth, the treatment of transformer layers and attention heads must be made explicit. Multi-head attention creates multiple parallel attention maps, and different heads may encode different relationships. A method may average heads before attribution, compute relevance separately for each head and average afterward, select a single head, or use a learned or gradient-based head weighting. Similarly, the selected transformer layer can substantially change the resulting heatmap. Authors should therefore report both the layer and the head aggregation strategy used to generate the final visualization.

Finally, the chosen representation should be appropriate for the claim being made. If a paper claims that Grad-CAM localizes image evidence, then the selected tensor should preserve a defensible relationship to image regions. If a paper uses cross-attention, the explanation should be framed as text-conditioned image-token attribution rather than as a generic visual saliency map. If a paper uses token embeddings, residual-stream activations, or MLP outputs, it should explain how spatial token identity is preserved or reconstructed.

Together, these requirements show that Grad-CAM adaptation for ViTs is not defined by a single formula, but by a set of architectural and implementation choices that must be reported explicitly. The goal is not to force all papers to use the same formulation, but to ensure that each visualization is specified precisely enough to be reproduced, compared, and interpreted.

Of course, not every paper has the space to derive a ViT-specific Grad-CAM formulation from first principles. However, citation cannot replace methodological specification. A sufficient citation should not merely point to the original CNN-based Grad-CAM paper or to an implementation repository, but to a work that defines the transformer feature representation, gradient target, spatial reconstruction procedure, and aggregation strategy being used. When authors rely on an existing ViT Grad-CAM adaptation, they should state how their implementation follows, modifies, or departs from that method. Additional reporting recommendations for authors and reviewers are provided in the Supplementary Materials (see~\ref{supp:recommendations}).

\section{Conclusion}\label{sec:conclusion}
Grad-CAM is widely used to explain ViT-based models, but its application to transformer architectures is often treated as more straightforward than it actually is. Unlike CNNs, ViTs do not naturally provide convolutional feature maps with fixed spatial and channel dimensions. Instead, visual information may be represented through patch tokens, attention matrices, residual-stream activations, MLP outputs, and, in VLMs, cross-attention maps. As a result, applying Grad-CAM to ViTs requires methodological choices that are not present in the original CNN formulation.

This study examined how Grad-CAM and Grad-CAM-adjacent methods are used across 175 papers involving ViTs or ViT-based architectures. The results show that most papers do not provide a full mathematical or implementation-level account of how Grad-CAM is adapted to transformer representations. Many cite only the original CNN-based Grad-CAM paper, cite implementation repositories, or describe the method only at a high level. These reporting patterns make it difficult to determine which tensor is being visualized, how gradients are aggregated, how non-spatial tokens are handled, and how the final heatmap should be interpreted.

To address this gap, we proposed a descriptive taxonomy of Grad-CAM adaptation choices for ViTs. The taxonomy distinguishes between token-feature methods, attention-map methods, residual-stream and MLP feature choices, head-wise aggregation strategies, and cross-attention variants in VLMs. It is not intended to prescribe a single correct formulation or to prove that one feature location is universally most faithful. Rather, it provides a framework for naming and comparing methodological choices that are often left implicit. The qualitative examples further show that these choices can produce substantially different visualizations for the same model and input, reinforcing that Grad-CAM on ViTs is not a uniquely defined procedure unless the adaptation is specified.

The central conclusion of this study is that Grad-CAM on ViTs should not be treated as a trivial extension of CNN-based Grad-CAM. Authors should report the feature extraction location, gradient target, token handling, spatial reconstruction procedure, and aggregation strategy used to generate each visualization. Reviewers should likewise treat the phrase ``Grad-CAM on a ViT'' as underspecified unless these details are provided directly or clearly inherited from a cited ViT-specific adaptation. Making these choices explicit is essential for rigor, reproducibility, and trust, especially in settings where visual explanations are used to support claims about model behavior.

\putbib[bibliography]

\end{bibunit}



\begin{bibunit}[IEEEtran]

\prefixcitationlinks{supp.}

\setcounter{section}{0}
\setcounter{figure}{0}
\setcounter{table}{0}

\renewcommand{\thefigure}{S\arabic{figure}}
\renewcommand{\thetable}{S\arabic{table}}
\renewcommand{\theequation}{S\arabic{equation}}

\renewcommand{\thesection}{S-\Roman{section}}

\title{Supplementary Material for Grad-CAM for Vision Transformers: A Systematic Taxonomy and Audit of Methodological Ambiguity in Explainable AI}

\author{Name\,\orcidlink{0009-0000-9973-6007}, 
Name\,\orcidlink{0009-0002-0638-5637}, \textit{Member, IEEE,} 
Name\,\orcidlink{0000-0002-4392-4188}, \textit{Member, IEEE,} 
Name\,\orcidlink{0000-0003-4176-0236}
\textit{Senior, IEEE} 

}

\maketitle

\section{Overview}\label{overview}

This supplementary material provides additional details supporting the main paper. Section~\ref{supp:blip_cross_attention}  presents a BLIP-base cross-attention Grad-CAM~\cite{grad_cam} example that illustrates word-level and layer-level variation in multimodal attribution maps. Section~\ref{supp:coding_scheme} defines the coding scheme used for the literature audit and provides the full paper-level coding table. Section~\ref{supp:limitations} discusses corpus construction limitations and coding assumptions. Section~\ref{supp:paper_level_mapping} provides detailed mathematical mappings and paper-level analysis of the ViT Grad-CAM adaptations examined in depth. Section~\ref{supp:recommendations} provides expanded reporting recommendations for authors and reviewers.

\section{BLIP Cross-Attention Grad-CAM Example}\label{supp:blip_cross_attention}

\begin{landscape}

\begin{figure}[p]
    \centering
    \includegraphics[width=\linewidth,height=0.99\textheight,keepaspectratio]{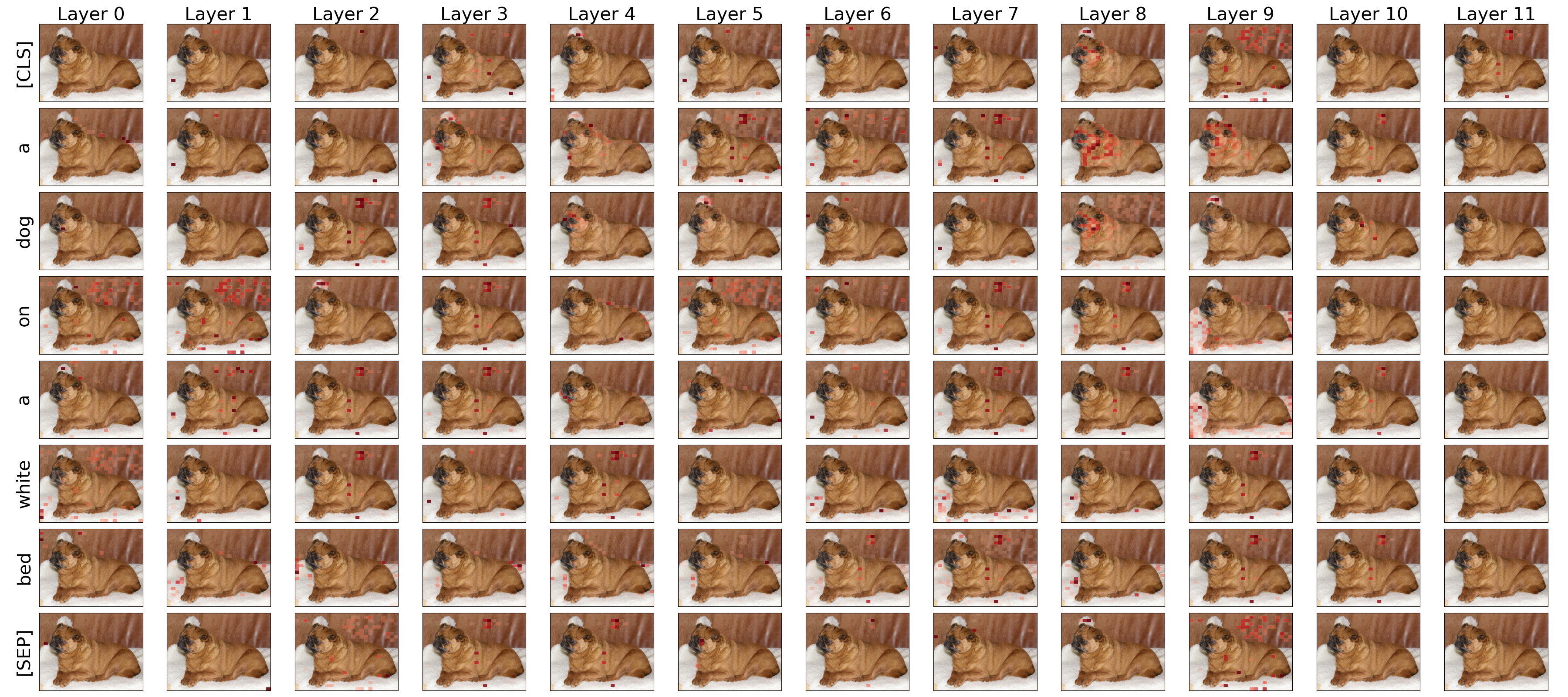}

    \caption{Cross-attention Grad-CAM visualizations for a BLIP model~\cite{blip} across all cross-attention layers for the caption ``A dog on a white bed.'' Rows correspond to cross-attention layers, and columns correspond to query tokens in the caption. Differences across heatmaps show that query-token choice and layer choice can substantially change the resulting localization pattern. Intermediate layers tend to produce more spatially localized maps, while later layers become more diffuse or vanish for non-\texttt{[CLS]} query tokens because their gradients approach zero.}
    \label{fig:blip_cross_attention_gradcam}
\end{figure}

\end{landscape}

Figure~\ref{fig:blip_cross_attention_gradcam} shows a concrete BLIP-base~\cite{blip} image-text matching example using the Hugging Face checkpoint \texttt{Salesforce/blip-itm-base-coco}, fine-tuned on COCO~\cite{coco}. We use the positive ITM logit as the target $y^c$ and compute separate cross-attention Grad-CAM maps for each query token across all cross-attention layers. The resulting maps illustrate that cross-attention Grad-CAM in VLMs is sensitive to both the selected query token and the selected layer: different words can produce different localization patterns, intermediate layers tend to produce more spatially localized maps, and later layers may become diffuse or vanish for non-\texttt{[CLS]} query tokens as their gradients approach zero.

\section{Corpus Coding Scheme and Full Coding Table}\label{supp:coding_scheme}

The supplementary coding table uses the following columns:
\begin{itemize}
\item \textbf{Citation}: Author names with a hyperlink to the corresponding reference.

\item \textbf{Year}: Year of publication.

\item \textbf{Venue}: Conference or journal in which the paper was published.

\item \textbf{Model Type}: Model architecture used in the paper, such as ViT, Swin Transformer, BEiT, CLIP, BLIP, or a related transformer-based vision architecture. We use the term ``ViT'' broadly to include models that rely fully or partially on tokenized image representations. When the exact architecture is not specified by the authors, we code the model type at the most specific level supported by the paper. We also indicate when a model is a hybrid architecture that combines convolutional and transformer components.

\item \textbf{Grad-CAM Role}: Role played by Grad-CAM in the paper. We use three categories: (1) ``Visualization'' when Grad-CAM is used for qualitative visualization, (2) ``Pipeline'' when Grad-CAM is used as a component of a method or pipeline, and (3) ``Analysis'' when Grad-CAM is used for quantitative analysis or evaluation.

\item \textbf{Cites Non-ViT Grad-CAM}: Indicates whether the paper cites prior work describing the original Grad-CAM method or a Grad-CAM variant originally designed for CNNs. A value of ``No'' means that the paper does not cite any prior Grad-CAM work, including the original Grad-CAM paper.

\item \textbf{Cites ViT Grad-CAM}: Indicates whether the paper cites prior work that applies Grad-CAM or a Grad-CAM-adjacent method to ViTs or ViT-based architectures. When such a citation is present, the cited paper is listed; otherwise, the entry is marked ``No.'' An asterisk (*) indicates that the paper mentions the cited work only in passing and does not clearly adopt or implement the method described in that work. The same notation is used in the Notes column when applicable.

\item \textbf{Task Type}: Computer vision or vision-language task for which Grad-CAM visualizations are produced. Common task types include classification, segmentation, weakly supervised semantic segmentation (WSSS), visual question answering (VQA), visual grounding, and other vision-language modeling (VLM) tasks.

\item \textbf{Notes}: Additional short notes about the paper, including links to sections where the paper is discussed in more detail.
\end{itemize}

\onecolumn

\makeatletter
\let\longtable@original@fnum@table\fnum@table
\renewcommand{\fnum@table}{\normalsize\longtable@original@fnum@table}
\makeatother
\tiny
\begin{longtable}{
  >{\raggedright\arraybackslash}p{0.17\textwidth} 
  >{\raggedright\arraybackslash}p{0.055\textwidth} 
  >{\raggedright\arraybackslash}p{0.02\textwidth} 
  >{\raggedright\arraybackslash}p{0.12\textwidth} 
  >{\raggedright\arraybackslash}p{0.08\textwidth} 
  >{\raggedright\arraybackslash}p{0.05\textwidth} 
  >{\raggedright\arraybackslash}p{0.09\textwidth} 
  >{\raggedright\arraybackslash}p{0.07\textwidth} 
  >{\raggedright\arraybackslash}p{0.13\textwidth} 
}
\caption{\normalsize Analysis of Grad-CAM Literature} \\
\hline
\textbf{Citation} & \textbf{Venue} & \textbf{Year} & \textbf{Model Type} & \textbf{Grad-CAM Role} & \textbf{Cites Non-ViT Grad-CAM} & \textbf{Cites ViT Grad-CAM} & \textbf{Task Type} & \textbf{Notes} \\ \hline
\endfirsthead

\multicolumn{9}{c}%
{{\normalsize \tablename\ \thetable{} -- continued from previous page}} \\[8pt]
\hline
\textbf{Citation} & \textbf{Venue} & \textbf{Year} & \textbf{Model Type} & \textbf{Grad-CAM Role} & \textbf{Cites Non-ViT Grad-CAM} & \textbf{Cites ViT Grad-CAM} & \textbf{Type} & \textbf{Notes} \\ \hline
\endhead

\hline \multicolumn{9}{r}{\normalsize{Continued on next page}} \\ \hline
\endfoot

\hline
\endlastfoot
H. Chefer et al.~\cite{Chefer_2021_CVPR} & CVPR & 2021 & ViT & Analysis & Yes & No & Classification & See~\ref{sec:chefer} \\
S. He et al.~\cite{He_2021_ICCV} & ICCV & 2021 & ViT & Visualization & Yes & No & Classification &  \\
H.   Chefer et al.~\cite{Chefer_2021_ICCV} & ICCV & 2021 & ViT & Analysis & Yes & \cite{Chefer_2021_CVPR}* & WSSS &  \\
B. Pan et   al.~\cite{NEURIPS2021_d072677d} & NeurIPS & 2021 & ViT & Visualization & Yes & \cite{Chefer_2021_CVPR}* &  \\
J.   Zhang et al.~\cite{NEURIPS2021_e02e27e0} & NeurIPS & 2021 & ViT & Visualization & Yes & No & Classification &  \\
T. Langlois et   al.~\cite{NEURIPS2021_e3603675} & NeurIPS & 2021 & ViT & Visualization & Yes & No & Classification &  \\
Y.   Xu et al.~\cite{NEURIPS2021_efb76cff} & NeurIPS & 2021 & Hybrid-ViT-CNN & Visualization & Yes & No & Classification &  \\
J. Li et   al.~\cite{NEURIPS2021_50525975} & NeurIPS & 2021 & ViT & Visualization & Yes & No & VQA & See~\ref{sec:align_before_fuse} \\
S.   He et al.~\cite{he2022vlmae} & arXiv & 2022 & VIT & Visualization & Yes & \cite{NEURIPS2021_50525975}* & VLM &  \\
M. Moayeri et   al.~\cite{Moayeri_2022_CVPR} & CVPR & 2022 & ViT & Visualization & Yes & No & Classification &  \\
H.   Li et al.~\cite{Li_2022_CVPR} & CVPR & 2022 & ViT & Visualization & Yes & No & Classification & pytorch-grad-cam \\
W. Yu et al.~\cite{Yu_2022_CVPR} & CVPR & 2022 & ViT & Visualization & Yes & No & Classification & Supp.   material \\
C.   Zhang et al.~\cite{Zhang_2022_CVPR} & CVPR & 2022 & ViT & Visualization & Yes & No & Classification &  \\
Y. Zheng et   al.~\cite{Zheng_2022_CVPR} & CVPR & 2022 & ViT & Visualization & Yes & No & Classification & possibly   pytorch-grad-cam \\
J.   Duan et al.~\cite{Duan_2022_CVPR} & CVPR & 2022 & ViT & Visualization & Yes & No & VLM &  \\
A. M. H. Tiong et   al.~\cite{tiong-etal-2022-plug} & EMNLP & 2022 & ViT & Pipeline & Yes & \cite{NEURIPS2021_50525975}* & VQA &  \\
C.   Si et al.~\cite{NEURIPS2022_94e85561} & NeurIPS & 2022 & ViT & Visualization & Yes & No & Classification &  \\
J. Guan et al.~\cite{NEURIPS2022_1d051fb6} & NeurIPS & 2022 & ViT & Visualization & Yes & No & Classification &  \\
H.   Choi et al.~\cite{Choi_2023_CVPR} & CVPR & 2023 & ViT & Analysis & Yes & \cite{Chefer_2021_CVPR} & Class Segmentation & See~\ref{sec:chefer} \\
H. Tang and K.   Jia.~\cite{Tang_2023_CVPR} & CVPR & 2023 & ViT & Visualization & Yes & No & Classification &  \\
X.   Fan et al.~\cite{Fan_2023_CVPR} & CVPR & 2023 & ViT & Visualization & Yes & No & Classification &  \\
Z. Cai et al.~\cite{Cai_2023_CVPR} & CVPR & 2023 & ViT & Visualization & Yes & No & Classification &  \\
C.   Oh et al.~\cite{Oh_2023_CVPR} & CVPR & 2023 & ViT & Visualization & Yes & No & Classification &  \\
C. Zhang et   al.~\cite{Zhang_2023_CVPR} & CVPR & 2023 & ViT & Visualization & Yes & \cite{tiong-etal-2022-plug}* & Classification &  \\
C.   Yu et al.~\cite{Yu_2023_CVPR} & CVPR & 2023 & ViT & Visualization & Yes & No & Classification &  \\
A. Nalmpantis et   al.~\cite{Nalmpantis_2023_CVPR} & CVPR & 2023 & ViT & Visualization & Yes & \cite{Chefer_2021_CVPR}* & Classification &  \\
G.   Zhang et al.~\cite{Zhang_2023_CVPR_2} & CVPR & 2023 & ViT & Visualization & No & No & Classification & No citation \\
T. Ronen et   al.~\cite{Ronen_2023_CVPR} & CVPR & 2023 & ViT & Pipeline & Yes & No & Classification &  \\
A.   Mohamed et al.~\cite{Mohamed_2023_CVPR} & CVPR & 2023 & ViT & Pipeline & Yes & \cite{Chefer_2021_CVPR} & Classification & See~\ref{sec:chefer} \\
K. Morrison et   al.~\cite{Morrison_2023_CVPR} & CVPR & 2023 & ViT & Analysis & Yes & No & Classification & pytorch-grad-cam \\
H.   Huang et al.~\cite{Huang_2023_CVPR} & CVPR & 2023 & Hybrid-ViT-CNN & Visualization & Yes & No & Classification & Unclear structure \\
F. Weers et   al.~\cite{Weers_2023_CVPR} & CVPR & 2023 & ViT & Visualization & Yes & \cite{NEURIPS2021_50525975}* & VLM &  \\
Y.   Chen et al.~\cite{Chen_2023_CVPR} & CVPR & 2023 & ViT & Visualization & Yes & \cite{NEURIPS2021_50525975}* & VLM &  \\
Y. Han et al.~\cite{Han_2023_CVPR} & CVPR & 2023 & ViT & Visualization & No & \cite{NEURIPS2021_50525975} & VLM & See~\ref{sec:align_before_fuse} \\
S.   Yu et al.~\cite{Yu_2023_CVPR_2} & CVPR & 2023 & ViT & Analysis & Yes & No & VLM &  \\
Z. Yang et al.~\cite{Yang_2023_CVPR} & CVPR & 2023 & ViT & Pipeline & Yes & \cite{NEURIPS2021_50525975} & VLM   / Grounding & See~\ref{sec:align_before_fuse} \\
Y.   Lin et al.~\cite{Lin_2023_CVPR} & CVPR & 2023 & ViT & Pipeline & Yes & No & VLM / Segmentation & See~\ref{sec:clip_segmenter} \\
J. Park and B.   Han.~\cite{Park_2023_CVPR} & CVPR & 2023 & ViT & Pipeline & Yes & \cite{NEURIPS2021_50525975} & VLM   / Segmentation & See~\ref{sec:align_before_fuse} \\
J.   Guo et al.~\cite{Guo_2023_CVPR} & CVPR & 2023 & ViT & Pipeline & Yes & \cite{NEURIPS2021_50525975}*   and~\cite{tiong-etal-2022-plug} & VQA & See~\ref{sec:plug_and_play_vqa} \\
J. Hanna et   al.~\cite{Hanna_2023_CVPR} & CVPR & 2023 & ViT & Visualization & Yes & No & WSSS &  \\
S.   Quan et al.~\cite{Quan_2023_ICCV} & ICCV & 2023 & ViT & Visualization & No & No & Classification & pytorch-grad-cam \\
M. Chen et al.~\cite{Chen_2023_ICCV} & ICCV & 2023 & ViT & Visualization & Yes & No & Classification &  \\
R.   Daroya et al.~\cite{Daroya_2023_ICCV} & ICCV & 2023 & ViT & Visualization & Yes & No & Classification &  \\
Y. Xu et al.~\cite{Xu_2023_ICCV} & ICCV & 2023 & ViT & Visualization & Yes & No & Classification &  \\
R.   Hesse et al.~\cite{Hesse_2023_ICCV} & ICCV & 2023 & ViT & Visualization & Yes & \cite{Chefer_2021_CVPR}* & Classification &  \\
S. He et al.~\cite{He_2023_ICCV} & ICCV & 2023 & ViT & Visualization & No & No & Regression & pytorch-grad-cam \\
B.   Shao et al.~\cite{Shao_2023_ICCV} & ICCV & 2023 & ViT & Visualization & No & \cite{NEURIPS2021_50525975}* & VLM & No direct citation  \\
J. Bi et al.~\cite{Bi_2023_ICCV} & ICCV & 2023 & ViT & Visualization & No & \cite{NEURIPS2021_50525975} & VLM & See~\ref{sec:align_before_fuse} \\
W.   Wang et al.~\cite{Wang_2023_ICCV} & ICCV & 2023 & ViT & Visualization & Yes & \cite{NEURIPS2021_50525975}* & VLM &  \\
G. Sun et al.~\cite{Sun_2023_ICCV} & ICCV & 2023 & ViT & Visualization & Yes & No & VLM &  \\
J.   Lee et al.~\cite{Lee_2023_ICCV} & ICCV & 2023 & ViT & Pipeline & Yes & \cite{NEURIPS2021_50525975} & VLM / Segmentation & See~\ref{sec:align_before_fuse} \\
D. Varam et al.~\cite{10262261} & IEEE   Access & 2023 & ViT & Visualization & Yes & No & Classification & pytorch-grad-cam \\
M.   Yao et al.~\cite{NEURIPS2023_ca0f5358} & NeurIPS & 2023 & ViT & Visualization & Yes & No & Classification & Supp. material \\
S. Tsutsui et   al.~\cite{NEURIPS2023_9f34484e} & NeurIPS & 2023 & ViT & Visualization & Yes & No & Classification &  \\
Y.   Lei et al.~\cite{NEURIPS2023_c2eac51b} & NeurIPS & 2023 & ViT & Visualization & Yes & No & Classification &  \\
Z. Wang et   al.~\cite{NEURIPS2023_2cf15395} & NeurIPS & 2023 & ViT & Visualization & Yes & \cite{NEURIPS2021_50525975} & VLM & See~\ref{sec:align_before_fuse} \\
Y.   Wang et al.~\cite{NEURIPS2023_339caf45} & NeurIPS & 2023 & ViT & Visualization & Yes & \cite{Chefer_2021_CVPR}* & VLM & pytorch-grad-cam. \\
Z. Wan et   al.~\cite{NEURIPS2023_af38fb8e} & NeurIPS & 2023 & ViT & Visualization & Yes & No & VLM &  \\
Y.   Zhao et al.~\cite{NEURIPS2023_a97b58c4} & NeurIPS & 2023 & ViT & Visualization & Yes & \cite{tiong-etal-2022-plug}* & VLM & pytorch-grad-cam.  \\
J. Xiao et al.~\cite{Xiao_2023_WACV} & WACV & 2023 & ViT & Analysis & Yes & \cite{Chefer_2021_CVPR}* & Classification & pytorch-grad-cam. \\
O.   Susladka et al.~\cite{Susladkar_2023_WACV} & WACV & 2023 & ViT & Visualization & No & \cite{NEURIPS2021_50525975}* & VLM & No direct citation \\
H. Shen et al.~\cite{Shen_2024_AAAI} & AAAI & 2024 & VIT & Pipeline & Yes & \cite{NEURIPS2021_50525975} & VLM & See~\ref{sec:align_before_fuse} \\
N.   BaoLong et al.~\cite{BaoLong_2024_ACCV} & ACCV & 2024 & ViT & Visualization & Yes & No & Classification &  \\
Y. Xiang et   al.~\cite{Xiang_2024_ACCV} & ACCV & 2024 & ViT & Visualization & No & No & Classification & No citation \\
M.   Jiang et al.~\cite{Jiang_2024_CVPR} & CVPR & 2024 & ViT & Analysis & Yes & No & Classification &  \\
D. Reilly and S.   Das.~\cite{Reilly_2024_CVPR} & CVPR & 2024 & ViT & Visualization & Yes & No & Classification &  \\
D.   Ye et al.~\cite{Ye_2024_CVPR} & CVPR & 2024 & ViT & Visualization & Yes & No & Classification &  \\
L. C. O. Tiong et   al.~\cite{Tiong_2024_CVPR} & CVPR & 2024 & ViT & Visualization & Yes & No & Classification &  \\
M.   Yang et al.~\cite{Yang_2024_CVPR} & CVPR & 2024 & ViT & Visualization & Yes & No & Classification &  \\
X. Tian et al.~\cite{Tian_2024_CVPR} & CVPR & 2024 & ViT & Visualization & Yes & No & Classification &  \\
F.   Parodi et al.~\cite{Parodi_2024_CVPR} & CVPR & 2024 & ViT & Visualization & Yes & No & Classification &  \\
Z. Lai et al.~\cite{Lai_2024_CVPR} & CVPR & 2024 & ViT & Visualization & Yes & No & Classification &  \\
Z.   Wang et al.~\cite{Wang_2024_CVPR} & CVPR & 2024 & Hybrid-ViT-CNN & Visualization & No & No & Classification & No citation \\
J. Yu et al.~\cite{Yu_2024_CVPR} & CVPR & 2024 & ViT & Visualization & Yes & No & Classification &  \\
T.   Mahmud et al.~\cite{Mahmud_2024_CVPR} & CVPR & 2024 & ViT & Visualization & Yes & No & Classification &  \\
Y. Luo et al.~\cite{Luo_2024_CVPR} & CVPR & 2024 & ViT & Visualization & Yes & No & Classification &  \\
W.   Zhu et al.~\cite{Zhu_2024_CVPR} & CVPR & 2024 & Swin & Visualization & Yes & No & Classification &  \\
W. Yu et al.~\cite{Yu_2024_CVPR_2} & CVPR & 2024 & ViT & Visualization & Yes & No & Classification & Supp.   material \\
H.   Karus et al.~\cite{Karus_2024_CVPR} & CVPR & 2024 & ViT & Visualization & Yes & No & Classification &  \\
B. H. Ngo et al.~\cite{Ngo_2024_CVPR} & CVPR & 2024 & ViT & Visualization & Yes & No & Classification &  \\
A.   Nasiri-Sarvi et al.~\cite{Nasiri-Sarvi_2024_CVPR} & CVPR & 2024 & ViT & Visualization & Yes & No & Classification &  \\
Y. Shang et   al.~\cite{Shang_2024_CVPR} & CVPR & 2024 & ViT & Visualization & Yes & No & Classification &  \\
A.   Althoupety et al.~\cite{Althoupety_2024_CVPR} & CVPR & 2024 & Hybrid-ViT-CNN & Visualization & Yes & No & Classification &  \\
F. Mehri et   al.~\cite{Mehri_2024_CVPR} & CVPR & 2024 & ViT & Analysis & Yes & \cite{NEURIPS2024_24f8dd1b}*   and~\cite{Chefer_2021_CVPR}* & Classification & pytorch-grad-cam. \\
J.   Wu et al.~\cite{Wu_2024_CVPR} & CVPR & 2024 & ViT & Analysis & Yes & \cite{Chefer_2021_CVPR} & Classification & See~\ref{sec:chefer} \\
K. Sumiyasu et   al.~\cite{Sumiyasu_2024_CVPR} & CVPR & 2024 & ViT & Analysis & Yes & \cite{Chefer_2021_CVPR}* & Classification & pytorch-grad-cam. \\
S.   B. Rongali et al.~\cite{Rongali_2024_CVPR} & CVPR & 2024 & ViT & Visualization & Yes & No & Clustering &  \\
Z. Zheng et   al.~\cite{Zheng_2024_CVPR} & CVPR & 2024 & ViT & Visualization & Yes & No & VLM &  \\
H.   Liu et al.~\cite{Liu_2024_CVPR} & CVPR & 2024 & ViT & Visualization & Yes & No & VLM &  \\
R. Ganz et   al.~\cite{Ganz_2024_CVPR} & CVPR & 2024 & ViT & Visualization & Yes & \cite{NEURIPS2021_50525975}* & VLM & \\
H.   Lin et al.~\cite{Lin_2024_CVPR} & CVPR & 2024 & ViT & Visualization & Yes & \cite{NEURIPS2021_50525975}* & VLM &  \\
Z. Hu et al.~\cite{Hu_2024_CVPR} & CVPR & 2024 & ViT & Pipeline & Yes & \cite{NEURIPS2021_50525975} & VLM & See~\ref{sec:align_before_fuse} \\
J.   Xie et al.~\cite{Xie_2024_CVPR} & CVPR & 2024 & ViT & Visualization & Yes & No & VLM & pytorch-grad-cam \\
Y. Zeng et al.~\cite{Zeng_2024_CVPR} & CVPR & 2024 & ViT & Pipeline & Yes & \cite{Lin_2023_CVPR}*   and~\cite{NEURIPS2021_50525975}* & VLM   / Grounding & See~\ref{sec:vlm_grounding} \\
J.   Luo et al.~\cite{Luo_2024_CVPR_2} & CVPR & 2024 & ViT & Pipeline & Yes & \cite{tiong-etal-2022-plug}*   and~\cite{NEURIPS2021_50525975}* & VLM / Segmentation & See~\ref{sec:emergent_segmentation}\\
S. Sun et al.~\cite{Sun_2024_CVPR} & CVPR & 2024 & ViT & Pipeline & Yes & \cite{Lin_2023_CVPR}* & VLM   / Segmentation & See~\ref{sec:clip_as_rnn} \\
R.   He et al.~\cite{He_2024_CVPR} & CVPR & 2024 & ViT & Pipeline & Yes & \cite{Chefer_2021_CVPR}*,~\cite{NEURIPS2021_50525975}*,   and \cite{tiong-etal-2022-plug} & VLM / Segmentation & See~\ref{sec:plug_and_play_vqa} \\
H. Zhang et al.~\cite{10706901} & IEEE   Access & 2024 & ViT & Visualization & Yes & No & Classification &  \\
J.   Aina et al.~\cite{10577973} & IEEE Access & 2024 & ViT & Visualization & Yes & No & Classification &  \\
Q. Zeng et   al.~\cite{NEURIPS2024_befcb9fa} & NeurIPS & 2024 & ViT & Visualization & Yes & No & Captioning &  \\
X.   Ma et al.~\cite{NEURIPS2024_f280a398} & NeurIPS & 2024 & ViT & Visualization & Yes & No & Classification &  \\
Z. Shu et   al.~\cite{NEURIPS2024_adb77ecc} & NeurIPS & 2024 & ViT & Pipeline & No & No & Classification & No citation \\
R.   Zeng et al.~\cite{NEURIPS2024_0a0eba34} & NeurIPS & 2024 & ViT & Visualization & Yes & No & Classification &  \\
W. Liu et   al.~\cite{NEURIPS2024_a5a5b0ff} & NeurIPS & 2024 & ViT & Visualization & Yes & No & Classification &  \\
J.   Teneggi and J. Sulam.~\cite{NEURIPS2024_8c1df815} & NeurIPS & 2024 & ViT & Visualization & Yes & No & Classification &  \\
S. Lin et   al.~\cite{NEURIPS2024_3a1fc7b8} & NeurIPS & 2024 & ViT & Visualization & Yes & No & Classification &  \\
Y.   Wang et al.~\cite{NEURIPS2024_d4ab6d24} & NeurIPS & 2024 & ViT & Visualization & Yes & \cite{Chefer_2021_CVPR}* & Classification & pytorch-grad-cam. \\
Y. Xu et   al.~\cite{NEURIPS2024_2d779258} & NeurIPS & 2024 & Swin & Visualization & Yes & No & Classification &  \\
S.   Balasubramanian et al.~\cite{NEURIPS2024_93e45db7} & NeurIPS & 2024 & ViT & Visualization & Yes & \cite{Chefer_2021_CVPR}* & Classification & \\
D. Ming et   al.~\cite{NEURIPS2024_24f8dd1b} & NeurIPS & 2024 & ViT & Pipeline & Yes & No & Classification & See~\ref{sec:boosting_transferability} \\
W.   Hu et al.~\cite{NEURIPS2024_59c147c7} & NeurIPS & 2024 & ViT & Visualization & Yes & No & LLM-like &  \\
C. Huang et   al.~\cite{NEURIPS2024_dda5cac5} & NeurIPS & 2024 & ViT & Visualization & Yes & No & Multi &  \\
M.   Dai et al.~\cite{NEURIPS2024_dc6319dd} & NeurIPS & 2024 & ViT & Visualization & Yes & No & VLM &  \\
H. Jung et   al.~\cite{NEURIPS2024_254404d5} & NeurIPS & 2024 & ViT & Visualization & Yes & No & VLM &  \\
T.   Li et al.~\cite{NEURIPS2024_4bbeef01} & NeurIPS & 2024 & ViT & Visualization & Yes & \cite{Chefer_2021_CVPR}* & VLM &  \\
G. Shen et   al.~\cite{NEURIPS2024_b1c62bde} & NeurIPS & 2024 & ViT & Visualization & Yes & No & VLM &  \\
L.   Yu et al.~\cite{NEURIPS2024_aec2dfc4} & NeurIPS & 2024 & ViT & Pipeline & Yes & \cite{NEURIPS2021_50525975}* & VLM & pytorch-grad-cam. \\
Y.-B. Lin et al.~\cite{10709650} & OJSC & 2024 & ViT & Visualization & Yes & No & Classification &  \\
D.   C. Bui et al.~\cite{Bui_2024_WACV} & WACV & 2024 & ViT & Visualization & Yes & No & Classification &  \\
S. Das et   al.~\cite{Das_2024_WACV} & WACV & 2024 & ViT & Visualization & Yes & No & Classification &  \\
P.   H. V. Valois et al.~\cite{Valois_2024_WACV} & WACV & 2024 & ViT & Analysis & Yes & \cite{Chefer_2021_CVPR}* & Classification &  \\
S. Black and R.   Souvenir~\cite{Black_2024_WACV} & WACV & 2024 & ViT & Visualization & Yes & No & Classification &  \\
R.   Ganz and M. Elad.~\cite{Ganz_2024_WACV} & WACV & 2024 & ViT & Visualization & Yes & No & VLM & Supp. material \\
X. Yang and X.   Gong.~\cite{Yang_2024_WACV} & WACV & 2024 & ViT & Pipeline & Yes & \cite{Lin_2023_CVPR} & VLM   / Segmentation & See~\ref{sec:clip_segmenter} \\
Y.   Wang et al.~\cite{Wang_AAAI_2025} & AAAI & 2025 & VIT & Pipeline & Yes & \cite{NEURIPS2021_50525975}*,~\cite{Sun_2024_CVPR}*,   and \cite{Luo_2024_CVPR_2}* & VLM & See~\ref{sec:iterprime} \\
F. Feng et al.~\cite{Feng_2025_CVPR} & CVPR & 2025 & ViT & Visualization & Yes & No & Classification &  \\
M.   Lou and Y. Yu.~\cite{Lou_2025_CVPR} & CVPR & 2025 & ViT & Visualization & Yes & No & Classification &  \\
G. Wang et al.~\cite{Wang_2025_CVPR} & CVPR & 2025 & ViT & Visualization & Yes & No & Classification &  \\
S.   Baek et al.~\cite{Baek_2025_CVPR} & CVPR & 2025 & ViT & Visualization & Yes & No & Classification &  \\
X. Li et al.~\cite{Li_2025_CVPR} & CVPR & 2025 & ViT & Visualization & Yes & No & Classification &  \\
S.   Alyami and H. Luqman.~\cite{Alyami_2025_CVPR} & CVPR & 2025 & ViT & Visualization & Yes & No & Classification &  \\
G. Jeannere et   al.~\cite{Jeanneret_2025_CVPR} & CVPR & 2025 & ViT & Visualization & Yes & \cite{Chefer_2021_CVPR}* & Classification & \\
N.   Echevarrieta-Catalan et al.~\cite{Echevarrieta-Catalan_2025_CVPR} & CVPR & 2025 & ViT & Analysis & Yes & \cite{Chefer_2021_CVPR}* & Classification & pytorch-grad-cam. \\
A. Chowdhury et   al.~\cite{Chowdhury_2025_CVPR} & CVPR & 2025 & ViT & Analysis & Yes & \cite{Chefer_2021_CVPR}* & Classification &  \\
H.   Zhong et al.~\cite{Zhong_2025_CVPR} & CVPR & 2025 & ViT & Visualization & No & No & Regression & pytorch-grad-cam \\
H. Choi et al.~\cite{Choi_2025_CVPR} & CVPR & 2025 & ViT & Visualization & Yes & No & VLM &  \\
Z.   Yang et al.~\cite{Yang_2025_CVPR} & CVPR & 2025 & ViT & Visualization & No & No & VLM & No citation \\
Y. Xie et al.~\cite{Xie_2025_CVPR} & CVPR & 2025 & ViT & Visualization & Yes & No & VLM &  \\
Y.   Zhang et al.~\cite{Zhang_2025_ICCV} & ICCV & 2025 & ViT & Visualization & Yes & No & Classification &  \\
A. Mehrpanah et   al.~\cite{Mehrpanah_2025_ICCV} & ICCV & 2025 & ViT & Analysis & Yes & No & Classification &  \\
J.   Yang et al.~\cite{Yang_2025_ICCV} & ICCV & 2025 & ViT & Visualization & Yes & No & Classification &  \\
Y. Cai et al.~\cite{Cai_2025_ICCV} & ICCV & 2025 & ViT & Visualization & Yes & No & Classification &  \\
R.   John et al.~\cite{John_2025_ICCV} & ICCV & 2025 & ViT & Visualization & Yes & No & Classification &  \\
X. Hao et al.~\cite{Hao_2025_ICCV} & ICCV & 2025 & ViT & Pipeline & Yes & No & Classification & See~\ref{sec:principles_visual_tokens} \\
Y.   Li et al.~\cite{Li_2025_ICCV} & ICCV & 2025 & ViT & Pipeline & Yes & \cite{Chefer_2021_CVPR}* & Classification &  \\
M. Kuroki et   al.~\cite{Kuroki_2025_ICCV} & ICCV & 2025 & ViT & Analysis & Yes & No & Classification & pytorch-grad-cam \\
Q.   Fan et al.~\cite{Fan_2025_ICCV} & ICCV & 2025 & ViT & Visualization & No & No & Clustering & No citation \\
H. Wang et al.~\cite{Wang_2025_ICCV} & ICCV & 2025 & ViT & Visualization & Yes & No & Regression &  \\
L.   Zhao et al.~\cite{Zhao_2025_ICCV} & ICCV & 2025 & ViT & Visualization & Yes & No & VLM &  \\
J. Tan et al.~\cite{Tan_2025_ICCV} & ICCV & 2025 & ViT & Visualization & Yes & No & VLM &  \\
A.   V. Aravindan et al.~\cite{Aravindan_2025_ICCV} & ICCV & 2025 & ViT & Analysis & Yes & \cite{Zeng_2024_CVPR}* & VLM & See~\ref{sec:vlm_bad_eyes} \\
X. Chen et al.~\cite{Chen_2025_ICCV} & ICCV & 2025 & ViT & Pipeline & Yes & \cite{NEURIPS2021_50525975}*   and~\cite{Lin_2023_CVPR}* & VLM   / Segmentation & See~\ref{sec:enhancing_prompt_generation} \\
T.   Hussain et al.~\cite{10938132} & IEEE Access & 2025 & Hybrid-CNN-ViT & Visualization & Yes & No & Classification &  \\
M. Yurdaku et al.~\cite{10975756} & IEEE   Access & 2025 & ViT & Visualization & Yes & No & Classification &  \\
K.   C. Pavithra et al.~\cite{11202415} & IEEE Access & 2025 & ViT & Visualization & No & No & Classification & No citation \\
X. Xue et al.~\cite{10839367} & IEEE   Access & 2025 & SE-ViT & Visualization & Yes & No & Classification & See~\ref{sec:sevit} \\
M.   Al-Imran et al.~\cite{11303744} & IEEE Access & 2025 & Hybid-Swin-CNN & Visualization & Yes & No & Classification &  \\
N. K. Jisy et al.~\cite{11230562} & IEEE   Access & 2025 & ViT & Visualization & Yes & No & Classification &  \\
P.   Bissoonauth-Daibo et al.~\cite{11186814} & IEEE Access & 2025 & ViT & Visualization & Yes & No & Classification &  \\
V. S. Mar´ın et al.~\cite{11426891} & IEEE   Access & 2025 & VIT-Quantum‑NAS-Hybrid & Visualization & No & No & Classification & No citation \\
T.   Sarveswaran and V. Rajangam~\cite{11218217} & IEEE Access & 2025 & Hybrid-ViT-CNN & Visualization & Yes & No & Classification &  \\
M. Khubaib et al.~\cite{11048568} & TCE & 2025 & Swin & Visualization & Yes & No & Classification &  \\
M.   R. A et al.~\cite{A_2025_WACV} & WACV & 2025 & ViT & Visualization & Yes & No & Classification &  \\
D. Opoku et al.~\cite{11328037} & IEEE   Access & 2026 & Hybid-Swin-CNN & Visualization & Yes & No & Classification &  \\
N.   Amin et al.~\cite{11373031} & IEEE Access & 2026 & Hybrid-ViT-CNN & Visualization & No & No & Classification & No citation \\
N. Penzel and J.   Denzle.~\cite{Penzel_2026_WACV} & WACV & 2026 & ViT & Analysis & Yes & No & Classification &  \\
N.-T.   Do-Tran et al.~\cite{Do-Tran_2026_WACV} & WACV & 2026 & ViT & Visualization & Yes & No & Classification &  \\
Y. Wu et al.~\cite{Wu_2026_WACV} & WACV & 2026 & ViT & Visualization & Yes & No & Classification &  \\
W.   Li et al.~\cite{Li_2026_WACV} & WACV & 2026 & ViT & Visualization & Yes & No & Classification &  \\
W. Li et al.~\cite{Li_2026_WACV_2} & WACV & 2026 & ViT & Visualization & Yes & No & Classification &  \\
E.   Chee et al.~\cite{Chee_2026_WACV} & WACV & 2026 & ViT & Pipeline & Yes & No & Classification &  \\
M. M. Rahman et   al.~\cite{Rahman_2026_WACV} & WACV & 2026 & ViT & Visualization & No & No & VLM & No citation \\
N.   Anand et al.~\cite{Anand_2026_WACV} & WACV & 2026 & ViT & Visualization & Yes & No & VLM &  \\
J. Cheng et   al.~\cite{Cheng_2026_WACV} & WACV & 2026 & ViT & Visualization & Yes & No & VLM &  \\
A.   Yadav et al.~\cite{Yadav_2026_WACV} & WACV & 2026 & ViT & Visualization & No & No & VLM & No citation\label{tab:all_papers}
\end{longtable}
\makeatletter
\let\fnum@table\longtable@original@fnum@table
\makeatother
\twocolumn
\normalsize
\twocolumn

\section{Corpus Construction Limitations and Coding Assumptions}\label{supp:limitations}

This literature audit should be understood as a systematic but bounded snapshot of a large, publicly accessible subset of the literature, rather than as a complete census of every paper that has applied Grad-CAM~\cite{grad_cam} to ViTs. Collecting papers for this task is inherently difficult because there is no standardized way to search for this specific methodological practice. Grad-CAM may appear in a title, abstract, method section, appendix, figure caption, or code repository, and authors do not always use consistent terminology when describing Grad-CAM or Grad-CAM-adjacent methods. Some papers mention Grad-CAM and ViTs in the introduction but never actually apply Grad-CAM to a ViT-based model, while others apply similar attribution methods without describing them in a way that is easy to retrieve through keyword search.

The initial search was conducted on April 17, 2026, using the query \texttt{("Grad-CAM" OR "Gradient-weighted Class Activation Mapping") AND ("Vision Transformer" OR "ViT")} for papers published from 2021 onward. Because the search was conducted before the end of 2026, the number of papers from 2026 should be interpreted as incomplete. Additional papers may have appeared after the search date, and some papers published near the time of the search may not yet have been indexed or made publicly available. For this reason, the publication-year analysis should be read as a snapshot of the literature at the time of collection rather than as a final measure of yearly publication trends.

The corpus was also shaped by access and venue constraints. After the initial keyword search returned a large number of candidate papers, we filtered the results to publicly available papers from the sources discussed in Section~III-B of the main paper.This decision was made so that the analyzed papers would be accessible to readers while also focusing on major computer vision, machine learning, and open-access publication venues. However, this filtering may exclude relevant work from other conferences, journals, workshops, preprint repositories, or domain-specific venues. As a result, the findings should not be interpreted as covering every possible use of Grad-CAM on ViTs, but rather as documenting patterns within a large, manually screened, and publicly accessible corpus.

The scope of the study also depends on broad definitions of both “Grad-CAM” and “ViT.” We use “Grad-CAM” to include Grad-CAM-adjacent methods such as Grad-CAM++~\cite{grad_campp}, LayerCAM~\cite{layer_cam}, ScoreCAM~\cite{score_cam}, ScoreCAM++~\cite{score_campp}, AblationCAM~\cite{ablation_cam}, FullGrad~\cite{full_grad}, and XGradCAM~\cite{axiom_grad_cam}. These methods were included because they share many of the same assumptions as Grad-CAM, especially the use of activation maps, gradients, scores, or feature-level attribution mechanisms originally designed around CNN-style representations. Similarly, we use “ViT” broadly to include architectures that rely fully or partially on token-based image representations, including standard ViTs, Swin Transformers~\cite{Liu_2021_ICCV}, BEiT~\cite{beit}, and ViT-based vision-language models such as CLIP~\cite{clip} and BLIP~\cite{blip}. This broader scope allows the analysis to capture a wider pattern of transformer-based attribution practices, but it also means that the included architectures and attribution methods are not identical.

Another limitation is that the coding process necessarily involved interpretive judgment. Many papers did not provide a complete mathematical specification of how Grad-CAM was applied to the transformer architecture. In these cases, we inferred the likely feature extraction location, gradient target, and aggregation strategy from the paper's textual description, architectural context, equations, figures, appendices, and cited methods. These inferences were made conservatively, and ambiguous cases were treated as ambiguous rather than forced into a single category. Even so, some categorizations may be debatable. This ambiguity is not merely a weakness of the present study; it is also one of the central findings. In many cases, the published description does not provide enough detail to determine exactly what “Grad-CAM on a ViT” means in the implementation.

This study also evaluates methodological specification rather than empirical faithfulness. We do not test whether each Grad-CAM visualization is faithful, stable, causally meaningful, or useful in downstream decision-making. A paper may provide a mathematically detailed Grad-CAM adaptation and still produce poor explanations, while another paper may provide limited methodological detail but produce visually plausible heatmaps. The focus here is narrower: whether the literature clearly specifies how Grad-CAM is adapted from CNNs to transformer-based architectures. The results should therefore be interpreted as a critique of methodological transparency and reproducibility, not as a direct benchmark of explanation quality.

Finally, the analysis is limited by what authors report in the papers themselves. Some authors may have used a well-defined implementation in code but not described it fully in the manuscript. Others may have relied on software packages, inherited code, or internal conventions without explicitly explaining those choices. However, from the perspective of scientific communication, this remains a limitation of the published work. If readers cannot determine the feature representation, gradient target, token handling, or aggregation procedure from the paper or its cited method, then the use of Grad-CAM is not fully reproducible from the publication alone.

Despite these limitations, the size of the screened corpus and the consistency of the observed patterns suggest that the central finding is robust: Grad-CAM is widely applied to ViTs and ViT-based architectures, but the mathematical and implementation details of this adaptation are often underspecified. The purpose of this meta-analysis is therefore not to provide a final exhaustive catalog of every use of Grad-CAM on ViTs, but to document a recurring methodological gap and provide a taxonomy for describing it more precisely.

\section{Paper-Level Mathematical Mapping and Analysis of ViT Grad-CAM Adaptations}
\label{supp:paper_level_mapping}

This section of the supplementary material provides a detailed, paper-level mapping of Grad-CAM adaptations to ViTs. Each subsection corresponds to a specific paper or set of papers that implement Grad-CAM or Grad-CAM-adjacent methods in the context of vision transformers. The analysis includes the feature representation used, the gradient target, the aggregation strategy, and any ambiguities or implementation details that are not fully specified in the original work. All references to the proposed taxonomy (e.g., $\mathcal{F}_1^{(l)}$, $\mathcal{F}_2^{(h,l)}$) are consistent with the definitions provided in the main paper.

\subsection{Grad-CAM for classification/segmentation (single-modal ViT)}\label{sec:gradcam_single_modal_vits}

\subsubsection{\texorpdfstring{Principles of Visual Tokens for Efficient Video Understanding~\cite{Hao_2025_ICCV}}{Principles of Visual Tokens for Efficient Video Understanding}}\label{sec:principles_visual_tokens}

\cite{Hao_2025_ICCV} uses a Grad-CAM-style oracle to estimate the importance of spatiotemporal tokens in a ViT-based video classification model. Unlike attention-map adaptations, the method operates on feature activations from the MLP in the final transformer blocks and uses the true-label class score as the gradient target. In the notation of the proposed taxonomy, this places the method closest to the token-activation branch, and more specifically closest to the MLP-output representation $\mathcal{F}_{5}^{(l)}$. If the implementation instead uses activations after the residual update following the MLP, the representation would be closer to $\mathcal{F}_{6}^{(l)}$, but the paper describes the source only as MLP feature activations.

Let $A^d_{thw}$ denote the feature activation for embedding dimension $d$ at temporal index $t$ and spatial position $(h,w)$. This can be interpreted as a reshaped token representation in which the token dimension has been arranged into a spatiotemporal grid, with $N = T \cdot H \cdot W$. Given the target class score $y^c$ before softmax, the method computes feature-importance weights by averaging gradients across all spatiotemporal token locations:

$$
\omega_d^c
=
\frac{1}{N}
\sum_t
\sum_h
\sum_w
\frac{\partial y^c}{\partial A^d_{thw}}.
$$
These weights are then used to compute a token-importance map through a Grad-CAM-style linear combination over the embedding dimension:
$$
S^c
=
\mathrm{ReLU}
\left(
\sum_d
\omega_d^c A^d
\right),
$$
where $S^c \in \mathbb{R}^{t \times h \times w}$ is the resulting token score map for class $c$. The method then applies min-max normalization so that the final patch scores lie between $0$ and $1$.

This formulation closely mirrors classical Grad-CAM. The embedding dimension $d$ plays the role of the CNN channel axis, while the reshaped spatiotemporal token grid $(T,H,W)$ serves as the analogue of the spatial feature map. The main difference is that the resulting map assigns importance to video tokens rather than image pixels or convolutional locations.

The paper does not fully specify whether $A$ is taken exactly from the MLP output or from a post-residual block representation after the MLP. Therefore, the safest taxonomy placement is $\mathcal{F}_{5}^{(l)}$, with a possible ambiguity toward $\mathcal{F}_{6}^{(l)}$ depending on implementation details. In either case, the method is best characterized as a token-activation Grad-CAM variant applied to spatiotemporal ViT features, using the true-label class logit as the gradient target.

Within the reviewed corpus, \cite{Hao_2025_ICCV} was not cited by any surveyed paper.

\subsubsection{\texorpdfstring{Transformer Interpretability Beyond Attention Visualization~\cite{Chefer_2021_CVPR}}{Transformer Interpretability Beyond Attention Visualization}}\label{sec:chefer} adapts Grad-CAM to ViTs by using the final attention layer rather than convolutional feature maps. The authors explicitly state that “the best way we found to apply GradCAM was to treat the last attention layer's [CLS] token as the designated feature map,” while excluding the \texttt{[CLS]} token itself. In the terminology of the proposed taxonomy, this places the method closest to a CLS-conditioned attention-map representation, corresponding most closely to $\mathcal{F}_2^{(h,l)}$.

The motivation for this choice comes from the structure of ViT classification. The final transformer output before the classification head has the form $v \in \mathbb{R}^{s \times d}$, where $s$ indexes tokens and $d$ indexes embedding dimensions. Since only the \texttt{[CLS]} token is passed to the classifier, applying Grad-CAM directly to $v$ produces sparse gradients: non-\texttt{[CLS]} tokens receive zero gradient because they do not directly determine the class score. To avoid this problem, the method instead uses the final attention layer's \texttt{[CLS]}-based representation as the attribution source.

This differs from token-feature Grad-CAM methods that treat the embedding dimension as the channel axis. Here, the attribution is tied to attention structure rather than to patch-token embeddings such as $\mathcal{F}_1$, $\mathcal{F}_4$, $\mathcal{F}_5$, or $\mathcal{F}_6$. The resulting visualization can be interpreted as a CLS-conditioned attention-based localization map, where relevance is derived from how the final attention layer connects the classification token to the remaining image tokens.

A key limitation is that the paper provides little formal justification for this design choice. The use of the final attention layer's \texttt{[CLS]} token is presented as a practical solution to the sparse-gradient problem rather than as a theoretically derived Grad-CAM analogue. As a result, some implementation details that are important in the proposed taxonomy, such as the exact attention tensor used, the treatment of individual heads, and the aggregation procedure, are not fully specified in the paper's description.

Overall, this method is best characterized as a final-layer, CLS-conditioned attention Grad-CAM variant. Its feature representation is closest to $\mathcal{F}_2^{(h,l)}$, but the paper's description is heuristic and leaves several taxonomy-relevant choices implicit.

Within the reviewed corpus, the Grad-CAM method from \cite{Chefer_2021_CVPR} is explicitly adopted by \cite{Choi_2023_CVPR,Mohamed_2023_CVPR,Wu_2024_CVPR} and mentioned in \cite{Chowdhury_2025_CVPR, Nalmpantis_2023_CVPR,NEURIPS2021_d072677d,Jeanneret_2025_CVPR, Chefer_2021_ICCV, Xiao_2023_WACV, Sumiyasu_2024_CVPR, Echevarrieta-Catalan_2025_CVPR, Valois_2024_WACV,Hesse_2023_ICCV, NEURIPS2024_93e45db7, NEURIPS2024_4bbeef01, Li_2025_ICCV, NEURIPS2023_339caf45, NEURIPS2024_d4ab6d24,Mehri_2024_CVPR, He_2024_CVPR}.


\subsubsection{\texorpdfstring{Squeeze-and-Excitation Vision Transformer (SE-ViT) for Lung Nodule Classification~\cite{10839367}}{Squeeze-and-Excitation Vision Transformer (SE-ViT) for Lung Nodule Classification}}\label{sec:sevit}

\cite{10839367} adapts Grad-CAM to SE-ViT by replacing CNN channels with image tokens. Rather than using an attention map as the attribution representation, the method operates on token-level feature activations. Let $A^x \in \mathbb{R}^{d}$ denote the feature vector associated with image token $x$, where $d$ is the token dimension. The authors do not specify the exact architectural source of $A^x$, so the representation cannot be assigned to a single feature extraction location. It is most consistent with a token-embedding representation such as $\mathcal{F}_1^{(l)}$, $\mathcal{F}_4^{(l)}$, $\mathcal{F}_5^{(l)}$, or $\mathcal{F}_6^{(l)}$, rather than an attention-map representation such as $\mathcal{F}_2^{(h,l)}$.

The main methodological change is the aggregation axis used to compute Grad-CAM weights. In standard CNN Grad-CAM, gradients are averaged over spatial locations to obtain one importance weight per channel. In SE-ViT, the authors instead compute one importance weight per image token by averaging gradients over the token dimension:
$$
w_x
=
\frac{1}{d}
\sum_i
\frac{\partial y}
{\partial A_i^x}.
$$
The resulting weights are therefore token-level weights rather than channel-level weights. The paper then forms its visualization map by combining token activations with these token weights:
$$
\mathrm{Map}
=
\mathrm{ReLU}
\left(
\sum_x
w_x A^x
\right).
$$

The authors explicitly state that the \texttt{[CLS]} token is removed during visualization because it is artificially added and does not correspond to image content. Therefore, the attribution is derived from image patch tokens only.

From the perspective of the taxonomy, this method is best characterized as a token-based Grad-CAM variant with token-wise gradient aggregation. However, several implementation details remain unclear. The paper does not identify the transformer block location from which $A^x$ is extracted, nor does it explain in detail how the weighted token activations are converted into the final spatial visualization. As a result, the method can be placed broadly within the token-embedding branch of the taxonomy, but it cannot be assigned precisely to a single feature representation such as $\mathcal{F}_1^{(l)}$, $\mathcal{F}_4^{(l)}$, $\mathcal{F}_5^{(l)}$, or $\mathcal{F}_6^{(l)}$.

Within the reviewed corpus, \cite{10839367} was not cited by any surveyed paper.

\subsubsection{\texorpdfstring{Boosting the Transferability of Adversarial Attack on Vision Transformer with Adaptive Token Tuning~\cite{NEURIPS2024_24f8dd1b}}{Boosting the Transferability of Adversarial Attack on Vision Transformer with Adaptive Token Tuning}}\label{sec:boosting_transferability}

\cite{NEURIPS2024_24f8dd1b} uses a Grad-CAM-inspired feature-importance computation to guide patch masking in a ViT. Rather than operating on attention maps, it fuses gradients and intermediate ViT features from a selected transformer layer. Let $F^{(l)}$ denote the feature representation extracted from layer $l$, and let $G^{(l)}$ denote the corresponding gradients. The paper defines the feature importance matrix as
$$
W
=
\sum_{i=1}^{C^{(l)}}
G_i^{(l)}
\odot
F_i^{(l)},
$$
where $C^{(l)}$ is the number of channels in layer $l$, and $\odot$ denotes element-wise multiplication.

In terms of the taxonomy, this places the method closest to the token-feature branch rather than the attention-map branch. The selected representation is an intermediate ViT feature tensor, so the channel dimension $C^{(l)}$ plays a role analogous to the embedding or feature dimension in token-based Grad-CAM adaptations. However, unlike classical Grad-CAM, the method does not explicitly compute channel weights by averaging gradients over spatial or token locations. Instead, it directly combines gradients and features element-wise and then sums over channels.

After constructing $W \in \mathbb{R}^{H \times W}$, the method partitions it according to the patch structure used for the sparse perturbation mask. This gives a patch-level representation $W_p = \{W^1_p,\dots,W^n_p\}$, where each $W^i_p \in \mathbb{R}^{P \times P}$ and $P$ represents the number of patches in the image. The importance of patch $x_p^{(i)}$ is then measured using the Frobenius norm: $\|W^i_p\|_F.$ These patch-level scores are min-max normalized and used to guide random patch dropping, so that patches with lower feature importance are more likely to be discarded while more important patches are more likely to be preserved.

The precise extraction location of $F^{(l)}$ is not specified. Therefore, it is unclear whether the method uses pre-attention token embeddings, post-attention representations, feedforward outputs, post-residual features, or another intermediate ViT tensor. In the notation used here, the representation could correspond to $\mathcal{F}_1^{(l)}$, $\mathcal{F}_4^{(l)}$, $\mathcal{F}_5^{(l)}$, $\mathcal{F}_6^{(l)}$, or another implementation-specific feature tensor.

The treatment of the \texttt{[CLS]} token is also not discussed. As a result, it is unclear whether the feature-importance computation uses all tokens, patch tokens only, or a representation from which non-spatial tokens have already been removed. The method can therefore be broadly characterized as a token-feature, Grad-CAM-inspired importance computation with patch-level aggregation, but several taxonomy-relevant implementation details remain unspecified.

Within the reviewed corpus, \cite{NEURIPS2024_24f8dd1b} was mentioned but not explicitly adopted as a Grad-CAM methodology by any surveyed paper.

\subsection{Grad-CAM for multi-modal ViTs}\label{sec:gradcam_multi_modal_vits}

\subsubsection{Cross-Attention Grad-CAM}

\mbox{}

\textbf{Plug-and-Play VQA: Zero-shot VQA by Conjoining Large Pretrained Models~\cite{tiong-etal-2022-plug}}\label{sec:plug_and_play_vqa} uses cross-attention maps as the attribution-bearing representation. Rather than extracting token embeddings such as $\mathcal{F}_1$, $\mathcal{F}_4$, $\mathcal{F}_5$, or $\mathcal{F}_6$, the method operates directly on a cross-modal attention matrix, corresponding to the $\mathcal{F}_2^{(h,l)}$ branch of the taxonomy.

Given image tokens $X \in \mathbb{R}^{K \times D_v}$ and text tokens $Y \in \mathbb{R}^{M \times D_t}$, the method defines a cross-attention map for each attention head. The cross-attention scores are written as
$$
A
=
\text{softmax}
\left(
\frac{Y W_Q W_K^\top X^\top}
{\sqrt{D_t}}
\right)
\in
\mathbb{R}^{M \times K},
$$
where the $j$th row indicates how much the $j$th text token attends to the $K$ image patches. In the taxonomy, this is a cross-attention variant of $\mathcal{F}_2^{(h,l)}$, since the representation explicitly encodes interactions between language tokens and visual tokens.

The method then computes the derivative of the similarity score with respect to the cross-attention scores and applies a Grad-CAM-style weighting directly to the attention entries. The relevance of image patch $i$ is defined as

$$
\text{rel}(i)
=
\frac{1}{H}
\sum_{j=1}^{M}
\sum_{h=1}^{H}
\max
\left(
0,
\frac{\partial sim(v,t)}
{\partial A^{(h)}_{ji}}
\right)
A^{(h)}_{ji}.
$$
This differs from conventional Grad-CAM because the method does not compute channel-level weights through spatial averaging. Instead, it multiplies each attention coefficient by the positive part of its corresponding gradient and then aggregates across text tokens and attention heads. Attribution is therefore computed directly at the level of cross-attention entries.

The resulting relevance score $\text{rel}(i)$ is associated with image patch $i$, so it can be mapped back to patch space for visualization or used to guide patch selection. Because attribution is computed over cross-attention from text tokens to image patches, the resulting explanation is explicitly multimodal and reflects image regions that contribute to image-text alignment.

Overall, this method is a clear example of cross-attention Grad-CAM in a multimodal ViT. Its feature extraction location is $\mathcal{F}_2^{(h,l)}$, its gradient target is the image-text similarity score $sim(v,t)$, and its aggregation procedure combines relevance across both text tokens and attention heads after element-wise gradient-attention weighting. One remaining ambiguity is that the exact selected layer of the Image-grounded Text Encoder (ITE) network is not specified beyond being a chosen cross-attention layer.

Within the reviewed corpus, the Grad-CAM method from \cite{tiong-etal-2022-plug} is explicitly adopted by \cite{Guo_2023_CVPR,He_2024_CVPR} and mentioned in \cite{Zhang_2023_CVPR,NEURIPS2023_a97b58c4,Luo_2024_CVPR_2}.


\textbf{Align before Fuse: Vision and Language Representation Learning with Momentum Distillation~\cite{NEURIPS2021_50525975}}\label{sec:align_before_fuse} is best characterized as an attention-map-based attribution method. Rather than using token-embedding representations such as $\mathcal{F}_1$, $\mathcal{F}_4$, $\mathcal{F}_5$, or $\mathcal{F}_6$, the method applies Grad-CAM-style localization to attention maps. In the notation used throughout this taxonomy, the relevant representation is therefore $\mathcal{F}_2^{(h,l)}$.

The paper considers two attribution settings. In the image-text contrastive (ITC) setting, Grad-CAM is computed on self-attention maps from the last layer of the visual encoder. The gradients are obtained by maximizing the image-text similarity score $s_{\text{itc}}$. During inference, the method computes importance scores for each image patch using self-attention with respect to the \texttt{[CLS]} token, then averages the resulting heatmaps across attention heads. This corresponds to a CLS-conditioned self-attention variant of $\mathcal{F}_2^{(h,l)}$.

In the ITM setting, Grad-CAM is instead computed on cross-attention maps from the 3rd layer of the multimodal encoder. The gradients are obtained by maximizing the ITM score $s_{\text{itm}}$. Here, the selected representation again corresponds to $\mathcal{F}_2^{(h,l)}$, but in a cross-attention setting where the attention maps encode interactions between text tokens and image tokens. The resulting scores are averaged across both attention heads and input text tokens to produce image-patch relevance scores.

This makes the distinction between the two variants clear. ITC performs attribution within the visual encoder using final-layer CLS-conditioned self-attention, while ITM performs attribution within the multimodal encoder using cross-attention after image and text information have been fused. The paper reports that the multimodal encoder produces finer-grained grounding because it better models image-text interactions.

Overall, ALBEF provides a relatively clear example of an $\mathcal{F}_2^{(h,l)}$-based attribution method. Its feature representation is an attention map, its gradient target is either $s_{\text{itc}}$ or $s_{\text{itm}}$, and its aggregation strategy involves averaging over attention heads for ITC and over both attention heads and text tokens for ITM. The main correction is that the paper does not indicate that text-token aggregation occurs before Grad-CAM; rather, it describes averaging the resulting scores across text tokens and heads.

Within the reviewed corpus, the Grad-CAM method from \cite{NEURIPS2021_50525975} is explicitly adopted by \cite{Bi_2023_ICCV,wall2026winsor,Han_2023_CVPR,Shen_2024_AAAI,Lee_2023_ICCV,Park_2023_CVPR,wang2021explaining,nayyem2024bridging,wang2025explainability, wang2025explainability1,Hu_2024_CVPR,NEURIPS2023_2cf15395,Yang_2023_CVPR} and mentioned in \cite{Shao_2023_ICCV,Susladkar_2023_WACV,tiong-etal-2022-plug,Weers_2023_CVPR,Lin_2024_CVPR,uddin2025expert,wang2026expert,NEURIPS2024_aec2dfc4,Ganz_2024_CVPR,he2022vlmae,uddin2026learning,Wang_2023_ICCV,Chen_2023_CVPR,wang2025explainability,wang2025bridging,Zeng_2024_CVPR,Wang_AAAI_2025,Chen_2025_ICCV,Luo_2024_CVPR_2,Guo_2023_CVPR,He_2024_CVPR}.


\textbf{Enhancing Prompt Generation with Adaptive Refinement for Camouflaged Object Detection~\cite{Chen_2025_ICCV}}\label{sec:enhancing_prompt_generation} applies gradient-weighted cross-attention to generate an initial localization prompt for camouflaged object detection. Similar to ALBEF~\cite{NEURIPS2021_50525975}, the relevant representation is a cross-attention variant of $\mathcal{F}_2^{(h,l)}$, where attention entries encode interactions between noun tokens and image tokens rather than purely visual self-attention.

Given image embeddings $\mathrm{Emb_{B_{\text{img}}}}$ and noun text input $t_{\text{noun}}$, the multimodal encoder produces both a matching score $S$ and a set of cross-attention maps $A_{\text{cross}}$:

$$
(S, A_{\text{cross}})
=
E_{B_{\text{text}}}
\big(
(\mathrm{Emb_{B_{\text{img}}}}, t_{\text{noun}})
\big).
$$

Gradients of the score $S$ with respect to the cross-attention maps are then computed, yielding $G_{\text{cross}}$. The initial localization map is generated by element-wise gradient-attention weighting:
$$
P_{\text{init}}
=\text{Grad-CAM}=
\mathrm{Mean}_{\text{twice}}
\left(
G_{\text{cross}}
\odot
A_{\text{cross}}
\right),
$$
where $\odot$ denotes element-wise multiplication and $\mathrm{Mean}_{\text{twice}}(\cdot)$ performs two averaging steps to aggregate the influence of the text on the image under different attention heads.

This places the method in the $\mathcal{F}_2^{(h,l)}$ branch of the taxonomy. The gradient signal is taken with respect to the matching score, but instead of computing channel-wise importance weights, the method applies the gradients directly to the cross-attention tensor. Attribution is therefore computed over individual attention entries rather than through a conventional Grad-CAM linear combination of feature channels.

The aggregation strategy is also relatively explicit. Relevance is first estimated at the level of cross-attention entries and then averaged twice, yielding the final multimodal prompt $P_{\text{init}}$. Thus, the method is best viewed as a gradient-gated cross-attention attribution procedure whose output is used as an initial localization prompt for camouflaged object detection.

Overall, this method is closely related to the ALBEF~\cite{NEURIPS2021_50525975} attribution framework. Both approaches use cross-attention maps rather than token embeddings as the Grad-CAM feature representation, use gradients with respect to attention coefficients as the attribution signal, and aggregate relevance after element-wise gradient-attention weighting. The main distinction is that this method uses the resulting map as part of a prompt-generation and refinement pipeline rather than solely as a post hoc visualization.

Within the reviewed corpus, \cite{Chen_2025_ICCV} was not cited by any surveyed paper.

\textbf{Iterative Grad-CAM Refinement for Referring Image Segmentation (IteRPrimE)~\cite{Wang_AAAI_2025}}\label{sec:iterprime} uses cross-attention activations as the basis for Grad-CAM-style attribution in referring image segmentation. The selected representation is closest to a cross-attention variant of $\mathcal{F}_2^{(h,l)}$, since the attention maps encode interactions between visual embeddings and textual tokens rather than token-embedding features such as $\mathcal{F}_1$, $\mathcal{F}_4$, $\mathcal{F}_5$, or $\mathcal{F}_6$.

Given an image-expression pair $(I,E)$, the visual and textual encoders produce embeddings $v$ and $e$, which are then fused through cross-attention layers. The resulting attention activation maps $A$ indicate which visual regions are activated with respect to each query token in the expression. Gradients are computed with respect to the selected attention representation using an output objective $y$, such as an ITM objective:
$$
G
=
\mathrm{clamp}
\left(
\frac{\partial y}{\partial A},
0,
\infty
\right),
$$
where negative gradients are removed. The Grad-CAM map is then formed by element-wise gradient-attention weighting:
$$
H
=
A
\odot
G,
$$
where $\odot$ denotes element-wise multiplication. This differs from classical Grad-CAM because the gradients are not pooled into channel-level weights. Instead, attribution is computed directly over the entries of the cross-attention activation map.

In the basic formulation described by the authors, the token-specific Grad-CAM maps are averaged over the expression tokens:
$$
H_f
=
\mathbb{E}_{k \in |e|}
\left[
H^k
\right],
$$
where $H_k$ denotes the Grad-CAM map for the $k$-th text token and $H_f \in \mathbb{R}^{B \times h \times w}$ is the resulting spatial map. This step shows that relevance is first computed at the level of individual text-token interactions and only then aggregated into an image-level localization map.

However, the paper identifies this simple averaging step as a limitation because it treats every word equally and ignores the relative importance of primary words in the referring expression. The proposed IteRPrimE framework then builds on this basic formulation by introducing word-aware aggregation and iterative refinement.

Overall, this method belongs to the gradient-gated cross-attention branch of the taxonomy. Its feature representation is closest to $\mathcal{F}_2^{(h,l)}$, its gradient target is an output matching or localization objective $y$, and its attribution mechanism is direct element-wise fusion of cross-attention activations with non-negative gradients. The important qualification is that the mean over text tokens describes the basic Grad-CAM formulation that IteRPrimE builds on, not necessarily the full final refinement strategy.

Within the reviewed corpus, \cite{Wang_AAAI_2025} was not cited by any surveyed paper.

\paragraph{Attention-Map Grad-CAM with Ambiguous Attention Source}

\mbox{}

\textbf{Emergent Open-Vocabulary Semantic Segmentation from Off-the-shelf Vision-Language Models~\cite{Luo_2024_CVPR_2}}\label{sec:emergent_segmentation} applies a Grad-CAM-style refinement to spatialized class attention maps in an ITM-based VLM setting. Unlike token-feature methods that operate on representations such as $\mathcal{F}_1$, $\mathcal{F}_4$, $\mathcal{F}_5$, or $\mathcal{F}_6$, this approach is closest to the $\mathcal{F}_2^{(h,l)}$ branch of the taxonomy. However, the paper treats the attention map as an already spatialized class-specific map rather than explicitly identifying a raw per-head attention matrix, layer, or token-level attention structure.

The method is applied in an ITM setting. Given an image-text pair, the model uses the ITM objective with the label ``matching'' to define the gradient signal. Let $M^{(k)} \in \mathbb{R}^{P \times P}$ denote the attention map for class $k$, and let $\mathcal{L}_{\text{ITM}}$ denote the corresponding ITM loss. The Grad-CAM-style salience map is computed as
$$
\mathrm{ReLU}
\left(
\frac{\partial \mathcal{L}_{\text{ITM}}}{\partial M^{(k)}}
\right)
\otimes
M^{(k)},
$$
where $\otimes$ denotes element-wise multiplication.

This formulation differs from classical Grad-CAM because it does not compute channel-wise importance weights through gradient pooling. Instead, the gradient of the ITM loss is applied directly to the attention map itself. The resulting map keeps attention scores whose gradients are positive under the matching-label objective and suppresses the remaining entries.

Several taxonomy-relevant details remain implicit. The paper does not clearly decompose $M^{(k)}$ by attention head, layer, text-token query, or image-token key. Instead, $M^{(k)}$ is treated as an already spatial attention map for class $k$. Therefore, any head aggregation, token aggregation, or spatialization step appears to occur before the reported Grad-CAM-style refinement or is not specified.

Overall, this method is best characterized as an attention-map refinement method closest to $\mathcal{F}_2^{(h,l)}$. Its gradient target is the ITM loss with the positive matching label, and its attribution mechanism is direct element-wise gradient-attention weighting. The main ambiguity is that the paper does not fully specify how the attention map $M^{(k)}$ is obtained or whether it corresponds to a particular layer, head, or token-level attention structure.

Within the reviewed corpus, \cite{Luo_2024_CVPR_2} was mentioned but not explicitly adopted as a Grad-CAM methodology by any surveyed paper.


\textbf{Investigating Compositional Challenges in Vision-Language Models for Visual Grounding~\cite{Zeng_2024_CVPR}}\label{sec:vlm_grounding} applies Grad-CAM-style attribution to an intermediate attention map in a vision-language transformer. The selected tensor, denoted $A_z$, is not described as a convolutional feature map or as a token-embedding representation such as $\mathcal{F}_1$, $\mathcal{F}_4$, $\mathcal{F}_5$, or $\mathcal{F}_6$. Instead, it is most consistent with the attention-map branch of the taxonomy, corresponding most closely to $\mathcal{F}_2^{(h,l)}$.

Given an image $\boldsymbol{v}$ and text query $\boldsymbol{t}$, the model extracts an intermediate attention representation from the multimodal transformer:

$$
A_z = f_z(\phi_f(\boldsymbol{v}, \boldsymbol{t})).
$$
The attribution target is a scalar image-text score $y$. Depending on the model, this score may correspond to image-text similarity under a contrastive objective or to an ITM score. Gradients are then computed with respect to the selected attention map:
$$
G_z = \frac{\partial y}{\partial A_z}.
$$
The grounding heatmap is formed by direct element-wise gradient-attention modulation:
$$
\mathrm{ReLU}
\left(
A_z
\odot
G_z
\right),
$$
where $\odot$ denotes element-wise multiplication. The resulting heatmap is resized to the input image resolution and used to identify image regions that explain the model's matching or similarity score.

This formulation differs from classical Grad-CAM because it does not compute channel-wise importance weights through spatial gradient averaging. Instead, the gradient signal is applied directly to the attention map itself. Attribution is therefore computed by gradient-gating attention activations rather than by constructing a weighted linear combination of feature channels.

Several implementation details remain implicit. The paper does not specify whether $A_z$ corresponds to self-attention, cross-attention, a particular head, or an aggregation over heads or tokens. Therefore, the method should be described as an attention-map Grad-CAM variant closest to $\mathcal{F}_2^{(h,l)}$, but not as a fully specified self-attention or cross-attention formulation.

Overall, this method is best characterized as a direct gradient-gated attention attribution approach. Its gradient target is an image-text compatibility score, its feature representation is an intermediate attention map, and its localization behavior depends on the selected layer $z$ within the multimodal transformer.

Within the reviewed corpus, \cite{Zeng_2024_CVPR} was mentioned but not explicitly adopted as a Grad-CAM methodology by any surveyed paper.

\paragraph{Vision-Encoder Token-Activation Grad-CAM in VLMs.}

\mbox{}

\textbf{CLIP as RNN: Segment Countless Visual Concepts without Training Endeavor~\cite{Sun_2024_CVPR}}\label{sec:clip_as_rnn} uses gradient-based CAM methods to generate mask proposals from a pre-trained CLIP model. Focusing specifically on the Grad-CAM integration described in the appendix, the method operates on feature maps from the ViT-based CLIP image encoder rather than on attention maps. It therefore does not correspond to $\mathcal{F}_2^{(h,l)}$. Instead, the paper states that gradients are computed using the feature map after the first normalization layer of the last residual block, which places the method closest to the layer-normalized token representation $\mathcal{F}_{(1,X)}^{(l)}$ .

Given an image $x$ and a set of text queries $h$, the image and text encoders produce feature vectors $v_x = f_I(x)$ and $v_h = f_T(h)$. The method computes an image-text similarity score using
$$
s
=
\mathrm{softmax}
\left(
v_x \cdot v_h^\top
\right),
$$
where the softmax is applied across the text-query dimension. This score serves as the attribution target. Gradients are then computed with respect to the selected image-encoder feature map $A^k$:

$$
g
=
\frac{\partial s}{\partial A^k}.
$$

The Grad-CAM weights are obtained by average-pooling these gradients over spatial locations:
$$
\alpha_k
=
\frac{1}{Z}
\sum_i
\sum_j
g^k_{ij},
$$
where $Z$ is the number of locations in each feature map. The final activation map is then computed using the standard Grad-CAM linear combination:
$$
L
=
\mathrm{ReLU}
\left(
\sum_k
\alpha_k A^k
\right).
$$

In the taxonomy, this is a token-activation Grad-CAM method rather than an attention-map method. The embedding dimension of the selected CLIP image-encoder feature map functions as the channel axis, and the resulting relevance map depends on reshaping the ViT feature representation into a spatial grid. Because the excerpt does not explicitly describe the reshape operation, the handling of the \texttt{[CLS]} token remains implicit, although it would need to be excluded or otherwise handled when producing a patch-level heatmap.

Overall, the Grad-CAM variant in this work is best characterized as an $\mathcal{F}_{(1,X)}^{(l)}$-style token-feature attribution method applied to CLIP. Its gradient target is a softmax-normalized image-text similarity score over text queries, and its feature extraction location is the first normalization layer of the last residual block of the CLIP image encoder. The broader mask proposal generator also uses CLIP-ES CAM~\cite{Lin_2023_CVPR} and class affinity modules, so this taxonomy placement applies specifically to the Grad-CAM component described in the implementation details.

Within the reviewed corpus, \cite{Sun_2024_CVPR} was mentioned but not explicitly adopted as a Grad-CAM methodology by any surveyed paper.


\textbf{CLIP is also an Efficient Segmenter: A Text-Driven Approach for Weakly Supervised Semantic Segmentation~\cite{Lin_2023_CVPR}}\label{sec:clip_segmenter} applies Grad-CAM to feature maps from the ViT-based CLIP image encoder in a weakly supervised semantic segmentation (WSSS) setting. The method does not use attention maps, so it does not correspond to $\mathcal{F}_2^{(h,l)}$. Instead, the paper states that Grad-CAM is computed using the feature map after the first normalization layer of the last residual block, placing the method closest to the token-embedding representation $\mathcal{F}_{(1,X)}^{(l)}$.

The method follows the standard Grad-CAM structure, but changes the score used to compute gradients. In ordinary Grad-CAM, the class weights are computed from the gradient of the class logit $Y^c$:
$$
w_k^c
=
\frac{1}{Z}
\sum_i
\sum_j
\frac{\partial Y^c}{\partial A_{ij}^k}.
$$
The class activation map is then formed as
$$
CAM_{ij}^c
=
\mathrm{ReLU}
\left(
\sum_k
w_k^c A_{ij}^k
\right).
$$

For CLIP, the paper introduces Softmax-GradCAM, where the target score is the softmax-normalized class/text score
$$
s^c
=
\frac{\exp(Y^c)}
{\sum_{c'=1}^{C}\exp(Y^{c'})}.
$$
Gradients are then computed using this softmax score rather than the raw class logit. As a result, the Grad-CAM weights incorporate competition between the target class and non-target classes:
$$
w_k^c
=
\frac{1}{Z}
\sum_i
\sum_j
\sum_{c'}
\frac{\partial Y^{c'}}
{\partial A_{ij}^k}
\frac{\partial s^c}
{\partial Y^{c'}}.
$$
Thus, Softmax-GradCAM does not change the selected feature representation, but it does change the gradient target and therefore the resulting channel weights. In the taxonomy, this belongs to the token-activation branch, with the ViT embedding dimension treated as the channel dimension and the patch-token structure reshaped into a spatial map for visualization.

An appendix of~\cite{Lin_2023_CVPR} further compares two ways of constructing the image-level representation used for Grad-CAM in ViT-based CLIP. One option uses the \texttt{[CLS]} token, while the other averages over the remaining patch tokens. The paper reports that average-token pooling produces more complete and accurate localization for WSSS, while \texttt{[CLS]}-based pooling tends to produce sharper but less spatially complete maps.

Overall, this method is best characterized as an $\mathcal{F}_1^{(l)}$-style token-feature Grad-CAM method applied to CLIP. Its feature extraction location is the first normalization layer of the last residual block, and its main Grad-CAM-specific modification is the use of a softmax-normalized image-text class score to compute gradient weights. The \texttt{[CLS]} versus average-token comparison affects the score construction and therefore the gradient signal, but not the underlying feature extraction location.

Within the reviewed corpus, the Grad-CAM method from \cite{Lin_2023_CVPR} is explicitly adopted by \cite{Yang_2024_WACV} and mentioned in \cite{Zeng_2024_CVPR, Sun_2024_CVPR, Chen_2025_ICCV}.


\textbf{Do VLMs Have Bad Eyes? Diagnosing Compositional Failures via Mechanistic Interpretability~\cite{Aravindan_2025_ICCV}}
\label{sec:vlm_bad_eyes} applies Grad-CAM-style attribution to intermediate activations from the vision component of a vision-language transformer. Unlike attention-map methods, it does not operate on $\mathcal{F}_2^{(h,l)}$. Instead, it treats a selected activation map from the ViT as the feature representation for Grad-CAM.

Given an image-text pair, the model computes a scalar similarity score $S_t$ measuring compatibility between the image and text prompt. Gradients of this score are then backpropagated into a chosen ViT layer, producing an activation tensor $
\mathbf{A} \in \mathbb{R}^{H \times W \times C},$ where $H$ and $W$ denote the spatial dimensions of the activation map and $C$ denotes the feature/channel dimension.

The method follows the standard Grad-CAM channel-weighting procedure. Channel importance weights are computed by averaging gradients over spatial locations:
$$
\alpha_c =
\frac{1}{H W}
\sum_{i=1}^{H}
\sum_{j=1}^{W}
\frac{\partial S_t}{\partial \mathbf{A}_{ij}^c}.
$$
The final saliency map is obtained by combining the activation channels with these weights:
$$
L_{\text{Grad-CAM}}
=
\mathrm{ReLU}
\left(
\sum_c
\alpha_c \mathbf{A}^c
\right).
$$

This places the method in the non-multi-headed activation-feature branch of the taxonomy. The selected representation is an activation tensor rather than a per-head attention map, so no head-level aggregation is required. The feature/channel dimension is treated as the analogue of the CNN channel axis, while the spatial dimensions of $\mathbf{A}$ provide the layout for the heatmap.

The exact extraction location of \(\mathbf{A}\) is not fully specified. If the activation is sampled before self-attention following layer normalization, it corresponds most closely to \(\mathcal{F}_{(1,X)}^{(l)}\). If captured after the attention mechanism, the feedforward network, or the residual updates, it would instead align with \(\mathcal{F}_{4}^{(l)}\), \(\mathcal{F}_{5}^{(l)}\), or \(\mathcal{F}_{6}^{(l)}\). However, the latter case is unlikely, as these later features typically cannot be simply reshaped into a spatial map. Despite this ambiguity, the method is clearly a token-feature or activation-feature Grad-CAM variant rather than an attention-map variant.

Within the reviewed corpus, \cite{Aravindan_2025_ICCV} was not cited by any surveyed paper.

\section{Reporting Recommendations for Authors and Reviewers}\label{supp:recommendations}

The findings of this study suggest that the use of Grad-CAM on ViTs would benefit from clearer reporting standards. The goal of these recommendations is not to discourage the use of Grad-CAM on transformer-based architectures, nor to require every paper to introduce a new explanation method. Rather, the goal is to ensure that when Grad-CAM is applied to ViTs, the adaptation is described precisely enough for readers to understand, reproduce, and evaluate the resulting heatmaps. Because ViTs differ from CNNs in how they represent spatial information, authors and reviewers should treat Grad-CAM on ViTs as a methodological choice that requires explicit specification rather than as a trivial extension of CNN-based Grad-CAM.

\subsection{For Authors}\label{sec:for_authors}

Authors who apply Grad-CAM or Grad-CAM-adjacent methods to ViTs should clearly identify the transformer representation used as the feature map. At minimum, papers should specify the selected layer, the type of representation used, and whether the attribution source is a token embedding, attention map, attention output, MLP activation, residual-stream representation, or cross-attention map. If the method uses a ViT representation as an analogue of a CNN activation map, the paper should explain how the token, channel, head, or spatial dimensions are being interpreted.

Authors should also describe how the heatmap is constructed from the selected representation. This includes specifying the gradient target, the treatment of non-spatial tokens such as \texttt{[CLS]}, the reshaping or interpolation procedure used to map token-level scores back to image space, and the aggregation strategy used across channels, heads, layers, or text tokens. These details are especially important in VLMs, where Grad-CAM may be computed with respect to image-text similarity, image-text matching, visual self-attention, or multimodal cross-attention.

When space is limited, authors should cite a prior work that fully specifies the Grad-CAM adaptation they are using. However, citing the original CNN-based Grad-CAM paper alone is not sufficient when the method is applied to a ViT. Similarly, citing an implementation repository is not a substitute for methodological explanation unless the relevant feature location, gradient target, and aggregation procedure are clearly described. If authors rely on a prior ViT-specific Grad-CAM adaptation that leaves some details ambiguous, they should state how their own implementation resolves those ambiguities.

Finally, authors should align their interpretation of the heatmap with the representation used to create it. A heatmap derived from patch-token embeddings should not be described in the same way as a heatmap derived from attention coefficients or cross-attention maps. Token-feature Grad-CAM, attention-map Grad-CAM, and cross-attention Grad-CAM can all produce spatial visualizations, but they do not necessarily explain the same computational object. Clear interpretation is therefore just as important as clear implementation.

\subsection{For Reviewers}\label{sec:for_reviewers}

Reviewers should treat the use of Grad-CAM on ViTs as a methodological claim that requires sufficient detail. If a paper states that it applies Grad-CAM to a ViT, reviewers should ask whether the authors specify the feature extraction location, gradient target, token handling, spatial reconstruction procedure, and aggregation strategy. Without these details, the visualization may not be reproducible, and readers may not be able to determine what the heatmap actually represents.

Reviewers should also be cautious when papers cite only the original Grad-CAM paper for a ViT-based application. Because Grad-CAM was originally designed for CNN feature maps, applying it to ViTs requires additional architectural assumptions. A paper does not need to provide an entirely new derivation, but it should either justify the adaptation directly or clearly cite a ViT-specific method that does so. If the cited method is itself underspecified, the authors should explain how their implementation fills in the missing details.

For papers in high-stakes domains, such as medical image analysis, reviewers should apply particular scrutiny to Grad-CAM visualizations used as evidence of model reliability or interpretability. In these settings, heatmaps may influence how readers understand or trust a model. If the attribution method is not clearly specified, the visualization may create an appearance of explanation without providing a reproducible account of the model's behavior. Reviewers should therefore ensure that claims based on Grad-CAM are proportional to the methodological detail provided.

Overall, reviewers can help improve the rigor of ViT interpretability research by requiring authors to distinguish between CNN-based Grad-CAM, token-feature Grad-CAM, attention-map Grad-CAM, and cross-attention Grad-CAM. These distinctions do not impose an unreasonable burden on authors; rather, they ensure that the reported visualizations are interpretable as scientific evidence rather than merely illustrative figures.

\onecolumn
\twocolumn



\renewcommand{\refname}{Supplementary References}
\putbib[bibliography]

\end{bibunit}

\end{document}